\documentclass[10pt,twocolumn,letterpaper]{article}

\usepackage{wacv}
\usepackage{times}
\usepackage{epsfig}
\usepackage{graphicx}
\usepackage{amsmath}
\usepackage{amssymb}
\usepackage{bbm}
\usepackage{rotating}
\usepackage{multirow}
\usepackage{multicol}
\usepackage[accsupp]{axessibility}
\usepackage{arydshln,enumitem}
\usepackage{float,nonfloat}
\usepackage{fancyhdr}

\def\wacvPaperID{1065 } % Enter the WACV Paper ID here

\wacvfinalcopy % *** Uncomment this line for the final submission

\ifwacvfinal
\usepackage[breaklinks=true,bookmarks=false]{hyperref}
\else
\usepackage[pagebackref=true,breaklinks=true,colorlinks,bookmarks=false]{hyperref}
\fi

\ifwacvfinal
\else
\fi

\begin{document}

%%%%%%%%% TITLE
\title{C-VTON: Context-Driven Image-Based Virtual Try-On Network%for Highly Realistic Human Body-Part Synthesis
}

\author{Benjamin Fele$^{1,2}$\vspace{-6mm} \\
%$^1$ Faculty of Electrical Engineering, $^2$ Faculty of Computer and information Sciences, University of Ljubljana, Slovenia\\
%{\tt\small {benjamin.fele, vitomir.struc}@fe.uni-lj.si, {ajda.lampe, peter.peer}@fri.uni-lj.si }
\and
Ajda Lampe$^{1,2}$\vspace{-6mm}\\
%University of Ljubljana, Slovenia\\
%{\tt\small ajda.lampe@fri.uni-lj.si}
\and
Peter Peer$^2$\vspace{-6mm}\\
%University of Ljubljana, Slovenia\\
%{\tt\small peter.peer@fri.uni-lj.si}
\and
Vitomir Štruc$^1$\vspace{-6mm}\\%\vspace{-7mm}
\and%\vspace{-7mm}
%{\vspace{-10mm}
$^1$Faculty of Electrical Engineering, $^2$Faculty of Computer and Information Science\\
University of Ljubljana, SI-1000 Ljubljana, Slovenia\\
{\tt\small \{benjamin.fele, vitomir.struc\}@fe.uni-lj.si, \{ajda.lampe, peter.peer\}@fri.uni-lj.si }
\vspace{40mm}}

\maketitle
\thispagestyle{fancy}
\fancyhead[C]{This paper was published at WACV 2022. Please cite the published version when referencing our work.}
%\begin{figure*}[!t!]
\noindent
\begin{minipage}[b]{\textwidth}
\vspace{-46mm}
\includegraphics[width=\textwidth]{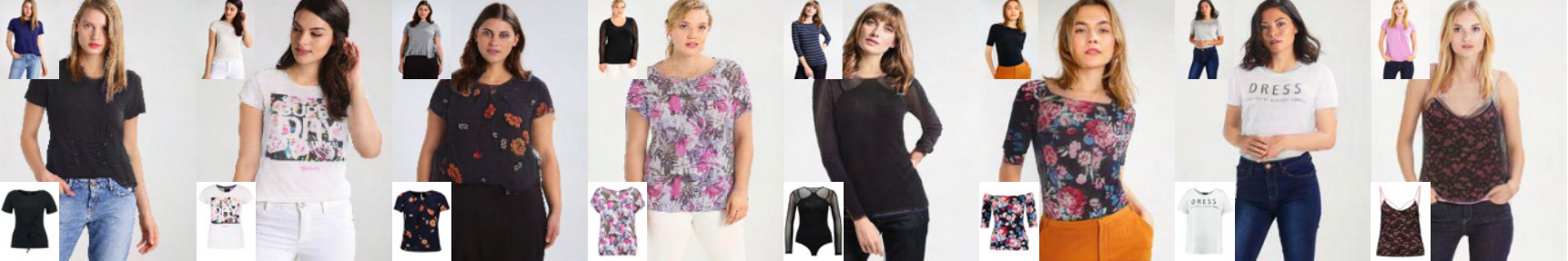}
\figcaption{Example results generated with the proposed Context-Driven Virtual Try-On Network (C-VTON). The original input image is shown in the upper left corner of each example and the target clothing in the lower left. Note that C-VTON generates convincing results even with subjects in difficult poses and realistically reconstructs on-shirt graphics.\vspace{9mm}}
\label{fig:banner}
\end{minipage}
%\end{figure*}

%%%%%%%%% ABSTRACT
\vspace{-3mm}
\begin{abstract}\vspace{-2mm}
   Image-based virtual try-on techniques have shown great promise for enhancing the user-experience and improving customer satisfaction on fashion-oriented e-commerce platforms. However, existing techniques are currently still limited in the quality of the try-on results they are able to produce from input images of diverse characteristics. In this work, we propose a Context-Driven Virtual Try-On Network (C-VTON) that addresses these limitations and  convincingly transfers selected clothing items to the target subjects even under challenging pose configurations and in the presence of self-occlusions. At the core of the C-VTON pipeline are: $(i)$ a geometric matching procedure that efficiently aligns the target clothing with the pose of the person in the input images, and $(ii)$ a powerful image generator that utilizes various types of contextual information when synthesizing the final try-on result. C-VTON is evaluated in rigorous experiments on the VITON and MPV datasets and in comparison to state-of-the-art techniques from the literature. Experimental results show that the proposed approach is able to produce photo-realistic and visually convincing results and significantly improves on the existing state-of-the-art. \vspace{-2mm}      
\end{abstract}

%%%%%%%%% BODY TEXT
\section{Introduction}

In the age of online shopping, virtual try-on technology is becoming increasingly important  and a considerable amount of research effort is being directed towards this area as a result~\cite{cheng2021fashion}. % and an increasing amount of research is conducted in this area. 
Especially appealing here are \textit{image-based} virtual try-on solutions that do not require specialized hardware and dedicated imaging equipment, but are applicable with standard intensity images, e.g.,~\cite{choi2021viton,dong2019towards,ge2021disentangled,ge2021parser,han2018viton}. As illustrated in Figure~\ref{fig:banner}, the goal of such solutions is to replace a piece of clothing in an input image with a target garment as realistically as possible. This allows for the design of virtual fitting rooms that let consumers try-on clothes without visiting (brick and mortar) stores, and also benefits retailers by   % on the consumer side, 
%helps to 
reducing product returns and  shipping costs % for businesses and, in turn, has positive effects on the environment by reducing carbon emission~
\cite{bertram2018study,cheng2021fashion,lee2020classification}.

Existing solutions to image-based virtual try-on are dominated by two-stage approaches (and their extensions) that typically include: $(i)$ \textit{a geometric matching} stage that aligns the target clothing to the pose of the person in the input image, and estimates an approximate position and shape of the target garment in the final try-on result, and $(ii)$ \textit{an image synthesis} stage that uses dedicated generative models (e.g., Generative Adversarial Networks (GANs)~\cite{goodfellow2014generative}), together with various refinement strategies to synthesize the final try-on image based on the aligned clothing and different auxiliary sources of information, e.g., pose keypoints, parsed body-parts, or clothing annotations among others \cite{choi2021viton, han2018viton, yang2020towards}. To further improve on this overall framework, various enhancements have also been proposed, including: $(i)$ refinements of the data fed to the geometric matching stage~\cite{minar2020cp, yang2020towards, yu2019vtnfp}, $(ii)$ integration of clothing segmentations into the synthesis procedure \cite{yang2020towards, yu2019vtnfp}, and $(iii)$ the use of knowledge distillation schemes to minimize the impact of %remove the dependence on human parsing models altogether
parser-related errors % on performance 
\cite{ge2021parser,issenhuth2020do}.

While the outlined advances greatly improved the quality of the generated try-on results, %several challenges still remain.  
%Aforementioned methods improve general quality of the generated virtual try-on, but still have some problems. Many approaches still suffer from 
the loss of details on the transferred garments that is often a consequence of difficulties with the geometric matching stage still represents a major challenge to image-based virtual try-on solutions~\cite{dong2019towards, han2018viton, wang2018toward}. Additionally, poor quality (human/clothing) parsing results typically still lead to unconvincing try-on images with garment textures being placed over incorrect body parts. Although recent (destilled) parser-free  models, e.g.,~\cite{issenhuth2020do,ge2021parser}, address this issue to some degree, they still inherit the main characteristics of the teacher models (including parsing issues) and often struggle with the generation of realistic body-parts, such as hands or arms.

In this paper, we propose a Context-Driven Virtual Try-On Network (C-VTON) that aims to address these issues. To improve the quality of the generated on-garment textures and logotypes, we design a novel geometric matching module that conditions the pose-matching procedure on body-segmentations only, and, therefore, minimizes the dependence on multiple (potentially error-prone) pre-processing steps. Additionally, we formulate learning objectives for the module's training procedure that penalize the appearance of the aligned clothing solely within the body area (while ignoring other body parts) to ensure that challenging pose configuration and self-occlusions (e.g. from hands) do not adversely affect performance. This design leads to realistic virtual try-on results with convincing details, as also shown in Figure~\ref{fig:banner}. Finally, we develop a powerful context-aware image generator (CAG) that utilizes contextual cues in addition to the warped clothing to steer the synthesis process. The generator is designed as a standard residual network, but relies on conditional (context-dependent) normalization operations (akin to SPADE layers~\cite{park2019semantic}) to ensure the contextual information is considered to a sufficient degree when generating virtual try-on result. In summary, we make the following main contributions in this paper:\vspace{-3mm}    % produce artefacts coming from the incapability of the model to match the pose of the person precisely. .  only within When learning the parameters   needed by most competing models. Because of strong structural priors associated with such body-part segmentations, this area can be estimated 

\begin{itemize}[leftmargin=*]
    \setlength\itemsep{-0.27em}
    \item We propose a novel image-based approach to virtual try-on, named Context-Driven Virutal Try-On Network (C-VTON), that produces state-of-the-art results with input images of diverse characteristics. %people wearing specified target garment as seen in Figure~\ref{fig:banner}. Specifically, we design an architecture that takes advantage of the Image Context (IC) consisting of masked person image, human body-part segmentations and original and warped target garment. Realistic image synthesis is achieved by utilizing IC in conditional normalization layers in the generator and multiple discriminators that assure realistic body-part generation,
    \item We design a simplified geometric matching module, termed Body-Part Geometric Matcher (BPGM), capable of producing accurate garment transformations even with subjects in challenging poses and arm configurations.
    \item We introduce a Context-Aware Generator (CAG) that allows for the synthesis of high-quality try-on results by making use of various sources of contextual information. %  Context-Aware Normalization (CAN) layer, building on SPADE~\cite{park2019semantic}, that conditionally normalizes layer activations based on Image Context (IC).
    \vspace{-2mm}
\end{itemize}
%-------------------------------------------------------------------------

\section{Related work}

% Do we need  a section for GANS?
%In this section we discuss work closely related to our proposed appraoch. For a more comprehensive coverage of the

%\textbf{Virtual Try-on.}  %Traditional approaches to \textit{Virtual Try-on} used computer graphics and relied on 3D models to render output images \cite{sekine2014virtual, ponsmoll2017clothcap}. Due to the 3D information and precise transformations, they allow for high-quality try-on result, however, their high computational complexity, as well as costly specialized hardware requirements stand in the way of widespread use of such systems.
%Due to advances in deep learning and (conditional) image synthesis, 
Image-based virtual try-on techniques have recently appeared as an appealing alternative to traditional try-on solutions that rely on 3D modeling and dedicated computer-graphics pipelines \cite{guan2012drape, patel2020tailornet,ponsmoll2017clothcap, santesteban2019learning}.  %modeling  on computationally intensive  have been showing promising performance with virtual try-on. 
The pioneering work from \cite{jetchev2017the, raj_eccv2018_swapnet}, for example, approached the virtual try-on task as an image analogy problem with promising results. However, due to the lack of explicit (clothing) deformation modelling, the generated images exhibited only limited photo realism. 
%such approaches fail to produce realistic results due to lack of explicit deformation modelling. % Furthermore, \cite{jetchev2017the} requires in-shop images of both source and target clothing, making it impractical for use in realistic scenarios. 
To address this shortcoming, Han \etal~\cite{han2018viton} proposed a two-stage approach, named VITON, that used a coarse-to-fine image generation strategy and utilized a Thin-Plate Spline (TPS) transformation~\cite{duchon1977splines} to align the image of the desired clothing with the pose of the target subject. %In the first stage, a coarse virtual try-on result is generated and in the second it is further refined  architecture corresponding to coarse and fine image generation. Additionally, they also use Thin-Plate Spline (TPS) transformations~\cite{duchon1977splines} to warp the clothing. 
Wang \etal~\cite{wang2018toward} improved on this approach with CP-VTON, which introduced a Geometric Matching Module (GMM) that allowed to learn the TPS clothing transformations in an end-to-end manner (similarly to~\cite{rocco2017convolutional}) and led to impressive try-on results. Follow-up work further refined the geometric matching stage using various mechanisms. CP-VTON+~\cite{minar2020cp}, for instance, improved the human mask fed to the GMM, VTNFP~\cite{yu2019vtnfp} designed an elaborate person representation as the input to the GMM, whereas LA-VITON~\cite{lee_iccvw2019_laviton} and ACGPN~\cite{yang2020towards} introduced additional transformation constraints when training their warping/matching modules. With C-VTON we follow the outlined body of work and design a novel matching module based on simplified inputs that can be estimated reliably even in the presence of considerable appearance variability. We achieve this by conditioning the module on body-parts only and leveraging the power of recent human-parsing models. % Because C-VTON avoids based on simplified inputs that can be estimated reliably using recent techniques. As we show in the experimental section, our design leads to highly competitive virtual try-on results, significantly outperforming the current state-of-the-art. %of our clothing transformation stage and carefully choose training objectives to match the body-area of the clothing as well as possible.

Several solutions have also been presented in the literature to improve the quality of the generated try-on results during the image synthesis stage. 
%Advancements are also being made in other aspects of the 
%in virtual try-on models. %To ensure higher quality of results at the refinement stage of image synthesis, 
MG-VTON~\cite{dong2019towards}, ACGPN~\cite{yang2020towards} and VITON-HD~\cite{choi2021viton}, for example, proposed using secondary neural networks that generate clothing segmentations matching the target garment and utilizing these as additional sources of information for the generator. S-WUTON~\cite{issenhuth2020do} and PF-AFN~\cite{ge2021parser} employed a teacher-student knowledge distillation scheme to alleviate the need for error-prone (intermediate) processing steps that often contribute to difficulties with existing try-on approaches. FE-GAN~\cite{dong2020fashion} and VITON-HD~\cite{choi2021viton} followed recent developments in image synthesis~\cite{park2019semantic} and introduced generators with conditional normalization layers to help with the quality and realism of the synthesized try-on results. Similarly to these techniques, C-VTON also uses an advanced image generator with conditional normalization layers in the synthesis step, but capitalizes on the use of contextual information to steer the  generation process. Furthermore, three powerful discriminators are employed in an adversarial training procedure to make full use of the available contextual information and improve the realism of the generated results.

\begin{figure*}[t]
\begin{center}
\includegraphics[width=0.85\linewidth]{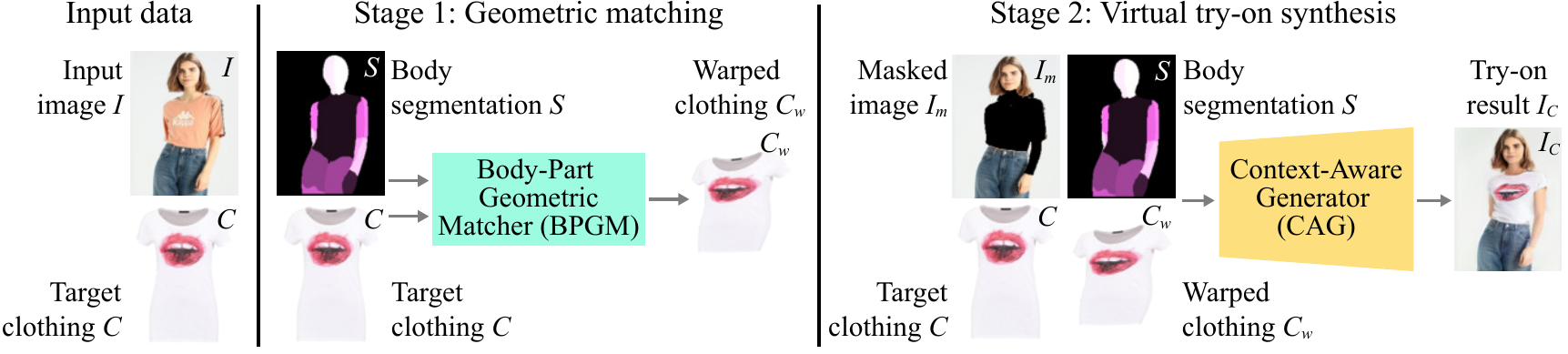}
\end{center}\vspace{-2mm}
\caption{Overview of the proposed Context-Driven Virtual Try-On Network (C-VTON) that given an input image of a subject $I$ and some target clothing $C$ generates a visually convincing virtual try-on image $I_C$. C-VTON is designed as a two-stage pipeline comprising a Body-Part Geometric Matcher (BPGM) that pre-aligns the target clothing $C$ with the pose of the subject in $I$ and a Context-Aware Generator (CAG) that generates the final try-on image $I_C$ based on the warped clothing and other sources of contextual information.} \vspace{-2mm}
\label{fig:arch}
\end{figure*}

\section{Context-Driven Virtual Try-On Network}\label{method}

%\begin{figure}
%\begin{center}
%\includegraphics[width=\linewidth]{fig/Legend.pdf}
%\end{center}\vspace{-3mm}
%   \caption{Image assets with corresponding names used in our pipeline.} \vspace{-3mm}
%\label{fig:legend}
%\end{figure}

% figure is placed right before related work for proper placement in the paper.

% \begin{figure*}
% \begin{center}
% \includegraphics[width=\linewidth]{fig/DPVTON_arch.pdf}
% \end{center}
%    \caption{Two-stage architecture of our proposed pipeline. a) Geometric matching module aimed to preserve target clothing patterns by matching them with original body-area texture. b) Context-Aware Generator and discriminators. Note that only the generator losses are disclosed in the figure.}
% \label{fig:arch}
% \end{figure*}

%\subsection{Overview of C-VTON}

We propose a Context-Driven Virtual Try-On Network (C-VTON) that relies on robust pose matching and contextualized synthesis to generate visually convincing virtual try-on results. Formally, given an input image of a subject, $I\in \mathbb{R}^{w \times h \times 3}$, and a reference image of some target clothing, $C \in \mathbb{R}^{w \times h \times 3}$, the goal of the model is to synthesize a photo-realistic output image, $I_{C}\in \mathbb{R}^{w \times h \times 3}$, of the subject from $I$ wearing the target clothing $C$ (see Figure \ref{fig:banner}).

As illustrated in Figure~\ref{fig:arch}, C-VTON is designed as a two-stage pipeline with two main components: $(i)$ a \textit{Body-Part Geometric Matcher (BPGM)} that warps the reference image of the target clothing $C$ to match the pose of the subject in the input image $I$, and $(ii)$ a \textit{Context-Aware Generator (CAG)} that uses the output of the BPGM together with various sources of contextual information to generate the final (virtual) try-on result $I_C$. Details on the two components are given in the following sections.

\subsection{The Body-Part Geometric Matcher (BPGM)} \label{sec:method:bpgm}

\begin{figure}[t]
\begin{center}
\includegraphics[width=\linewidth]{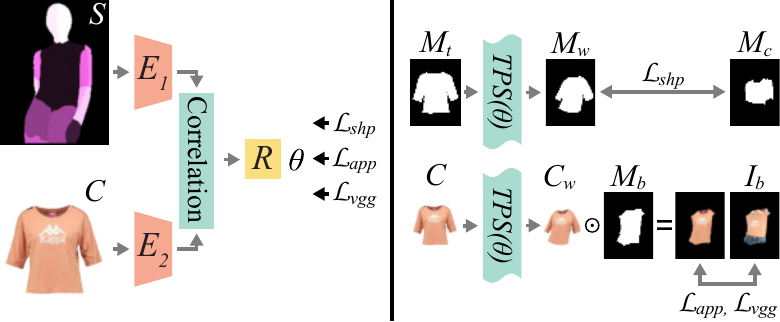}
\end{center}\vspace{-3mm}
   \caption{Overview of the Body-Part Geometric Matcher (BPGM). The BPGM architecture is shown on the left and the training losses designed to preserve on-garment textures and overall garment shape are on the right. Unlike competing solutions, BPGM estimates the warping function based on body-part locations only, leading to robust performance even in the presence of challenging poses, e.g., with crossed arms, arms occluding the body, etc.} \vspace{-3mm}
\label{fig:bpgm}
\end{figure}

The first stage of the C-VTON pipeline consists of the proposed BPGM,  and is responsible for estimating the parameters of a Thin-Plate Spline (TPS)~\cite{duchon1977splines} transformation that is used to align the target clothing with the pose of the person in the input image $I$. This allows for approximate positional matching of the target garment and helps to make the task of the generator in the next stage easier. As shown in Figure~\ref{fig:bpgm}, the BPGM  takes a reference image of the target clothing $C$ and body segmentations $S$ as input, and then produces a warped version $C_w$ of the target clothing at the output. The body segmentations $S\in \{0, 1 \}^{w \times h \times d}$ are generated using the DensePose model from \cite{Guler2018DensePose} and contain $d=25$ channels (classes), each corresponding to a different body part. Compared to other virtual try-on architectures that utilize complex clothing-agnostic person representations, e.g., %images and pose keypoints 
\cite{choi2021viton, issenhuth2020do, wang2018toward}, to obtain geometric clothing transformations, BPGM relies on body-part segmentations only, which are sufficient for reliably matching target garments to person images, as we show in our experiments. %Our approach alleviates the issues with transforming garments to people's poses that the Thin-Plate Spline (TPS)~\cite{duchon1977splines} transformations can't match. % and minimizes the influence of avoids using clothing segmentations which often contain noisy labels.

\textbf{BPGM Architecture.} The proposed BPGM follows the design of the Geometric Matching Module (GMM) from \cite{wang2018toward}  and consists of two distinct encoders, $E_1$ and $E_2$. The first takes the target clothing $C$ as input and generates a corresponding feature representation $\psi_{{E_1}}\in \mathbb{R}^{{w_f} \times {h_f} \times {d_f}}$. Similarly, the second encoder accepts the body segmentations $S$ at the input and produces a feature representation $\psi_{{E_2}}\in \mathbb{R}^{{w_f} \times {h_f} \times {d_f}}$ at the output. Here,  %The second encoder    and body segmentation $S$, respectively. First, we obtain the output features from each encoder $\Psi_{\{E_1, E_2\}} \in \mathbb{R}^{\Bar{w} \times \Bar{h} \times \Bar{c}}$, where 
$w_f$ and ${h_f}$ represent spatial dimensions that depend on the depth of the encoders and ${d_f}$ denotes the number of output channels. Next, the feature representations are normalized channel-wise to unit $L_2$ norm, spatially flattened and organized into a matrix ${\Psi}_{\{E_1, E_2\}}\in \mathbb{R}^{{d_f\times{w_f}{h_f}}}$, which %. This matrix finally 
serves as the basis for computing the correlation matrix $Corr$~\cite{rocco2017convolutional}, i.e.:
\begin{equation}
    Corr = {\Psi}_{E_1}^\top{\Psi}_{E_2}\in \mathbb{R}^{({w_f}{h_f}) \times ({w_f}{h_f})}.
\end{equation}
The correlation matrix $Corr$ is then fed into a regressor, $R$, which predicts a parameter vector $\theta$ (with $2n^2$ dimensions) that corresponds to $x$ and $y$ offsets on an $n \times n$ grid, according to which the target clothing $C$ is warped. 

\textbf{Training Objectives.}  Three loss functions are used to learn the parameters of the BPGM,, i.e.:
\begin{itemize}[leftmargin=*]
\setlength\itemsep{-0.1em}
    \item A \textbf{target shape loss} ($\mathcal{L}_{shp}$) that encourages the warping procedure to render the target clothing in a shape that matches the pose of the subject in $I$, i.e.:
    \begin{equation}
    \mathcal{L}_{shp} = \lVert M_{w} - M_{c} \rVert_1 = \lVert T_\theta(M_t) - M_{c} \rVert_1,
    \label{eq: shape loss}
    \end{equation}
    where $M_t$ and $M_w$ are binary masks corresponding to the original ($C$) and warped target clothing ($C_w$), respectively, $M_c$ is a binary mask corresponding to the clothing area in the input image (generated with the segmentation model of Li \etal from \cite{li2020self}), and $T_\theta$ denotes the TPS transformation parameterized by $\theta$.
    \item An \textbf{appearance loss} ($\mathcal{L}_{app}$) that forces the visual appearance of the warped clothing $C_w$ within the body area $M_b$ to be as similar as possible to the input image $I$, i.e.:  
    \begin{equation}
    \mathcal{L}_{app} = \lVert C_{w} \odot M_{b} - I_{b} \rVert_1,
    \label{eq: appearence loss}
    \end{equation}
    where $\odot$ is the Hadamard product and $M_b$ a binary masks of the body area (a channel in $S$), and $I_{b} = I\odot M_{b}$.
    \item A \textbf{perceptual loss} ($\mathcal{L}_{vgg}$) that ensures that the target clothing and its warped version contain the same semantic content  within the body area, i.e.~\cite{ledig2017photo}:
    \begin{equation} \label{eq:vggloss}
        \mathcal{L}_{vgg} = \sum_i^n \lambda_i \lVert \phi_i(C_{w} \odot M_{b}) - \phi_i(I\odot M_{b}) \rVert_1,
    \end{equation}
    where $\phi_i(\cdot)$ is a feature map generated before each (of the $n=5$) max-pooling layer of a VGG19~\cite{simonyan2014very} model (pretrained on ImageNet), and $\lambda_i$ is the corresponding weight. % for $i$-th feature distance, and the variables, $X' = C_{w} \times M_{b}$ and $X = I_{b}$. $\mathcal{L}_{mask}$
\end{itemize}
%We warp target clothing mask $M_{t}$ to obtain $M_{w}$ and $C_{t}$ to get $C_{w}$. BPGM's parameters are optimized using 3 loss terms; two $\mathcal{L}_1$ losses:
%\begin{align}
%    & \mathcal{L}_{shape} = \lVert M_{w} - M_{c} \rVert_1 \\
%    & \mathcal{L}_{appear} = \lVert (C_{w} \times M_{b}) - I_{b} \rVert_1
%\end{align}
%and a perceptual loss~\cite{ledig2017photo}:
%\begin{equation} \label{eq:vggloss}
%    \mathcal{L}_{VGG}(X', X) = \sum_i^5 \lambda_i \lVert \Phi_i(X') - \Phi_i(X) \rVert_1,
%\end{equation}
%where $\Phi_i(x)$ above is a feature map before each max-pooling layer in a VGG19~\cite{simonyan2014very} network pretrained on ImageNet and $\lambda_i$ is the corresponding weight for $i$-th $L_1$ feature distance. 
Among the above losses, $\mathcal{L}_{shp}$ aims to match the general garment area, while $\mathcal{L}_{app}$ and $\mathcal{L}_{vgg}$ are designed to specifically match the on-garment graphics, without forcing the BPGM matcher to align sleeves, which are % as well -- the latter is 
often a source of unrealistic transformations used later by the generator. 

Finally, the \textbf{joint learning objective}  for the BPGM is: 
\begin{equation} \label{eq:bpgmjoint}
%\begin{aligned}
    \mathcal{L}_{BPGM}  = \lambda_{shp} \mathcal{L}_{shp} + \lambda_{app} \mathcal{L}_{app} 
                        + \lambda_{vgg} \mathcal{L}_{vgg},
%\end{aligned}
\end{equation}
where $\lambda_{shp}$, $\lambda_{app}$ and $\lambda_{vgg}$ are balancing weights. The parameters of the BPGM are learned over a dataset of input images $I$ with matched images of target clothing $C$. % We note that the training procedure requires a reference image of the clothing $C$ already present in $I$ (available in relevant datasets) to be able to learn the parameters of the BPGM. %- this is not needed at run-time. 

\subsection{The Context-Aware Generator (CAG)} \label{sec:method:cag}

\begin{figure}[t]% \subsection{Architecture (take out / restructure)}
\begin{minipage}[b]{0.9\columnwidth}
\begin{center}
% \fbox{\rule{0pt}{2in} \rule{0.96\linewidth}{0pt}}
\includegraphics[width=0.97\linewidth, trim = 0 32mm 0 4mm, clip]{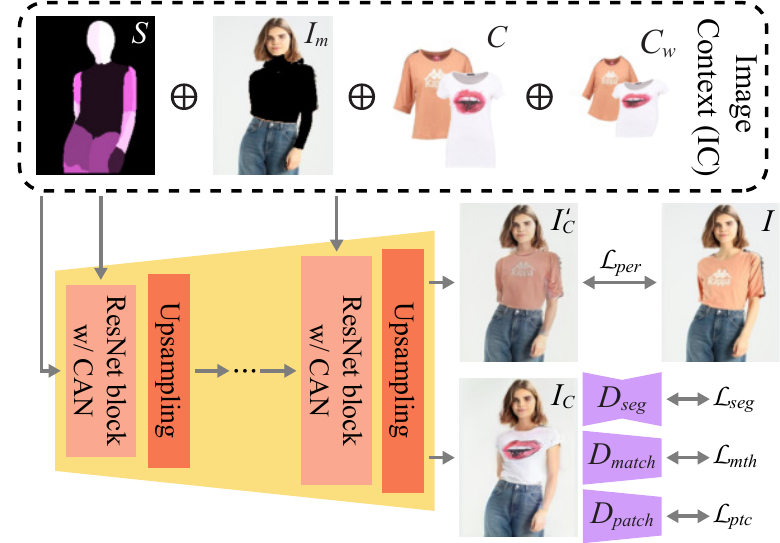}\vspace{1mm}
\small (a) Context-Aware Generator (CAG)
\end{center}
\end{minipage}\\[2mm]
\begin{minipage}[b]{0.9\columnwidth}
\begin{center}
\includegraphics[width=0.94\linewidth, trim = 0 0 0 63.5mm, clip]{fig/CAG_arch_v2.pdf}\vspace{1mm}
\small (b) ResNet block w/ Context-Aware Normalization (CAN)\vspace{2.5mm}

\end{center}
\end{minipage}\vspace{-1mm}
   \caption{Overview of the Context-Aware Generator (CAG): (a) the proposed generator with associated losses used during training, (b) a schematic representations of a ResNet block with Context-Aware Normalization (CAN). The proposed CAG is designed to exploit contextual information for the synthesis step - at the input, for activation normalization and through a series of discriminators. \vspace{-5mm}} % a) During training, CAG receives context $IC$ with either a matching or non-matching target garment producing outputs optimized using multiple losses. b) CAN layer is positioned before convolutional layers in ResNet blocks, aimed at ensuring spatially dependant activation normalization. \vspace{-2mm}}
\label{fig:cag}
\end{figure}

The second stage of the C-VTON pipeline consists of the Context-Aware Generator (CAG) and is responsible for synthesizing the final virtual try-on image $I_C$ given the warped target clothing and other contextual cues as input. To simplify the discussion in the remainder of the section, we jointly refer to all inputs of the generator as \textit{Image Context (IC)} hereafter, and define it as a channel-wise concatenation of the body-part segmentations $S$, the input image with masked clothing area $I_m$ (computed as $I_m = I\odot M_c$), and the target and warped clothing images, $C$ and $C_w$, respectively, i.e., $IC = S \oplus I_m \oplus C \oplus C_w$. A visual illustration of $IC$ is presented at the top of Figure~\ref{fig:cag}(a).        

\textbf{CAG Design.} The context-aware generator consists of a sequence of  ResNet blocks~\cite{he2016deep} and ($2\times$) upsampling layers augmented with what we refer to as \textit{Context-Aware Normalization (CAN)} operations. Similarly to recent spatially-adaptive normalization mechanisms used in the field of conditional image synthesis~\cite{park2019semantic, schonfeld2020you}, the proposed CAN layers are designed to efficiently utilize the information from the image context $IC$ and feed the generator with critical contextual information. As illustrated Figure~\ref{fig:cag}(a), this is done %on the desired semantic layout of the output image 
at different resolutions to ensure $(i)$ that the activations of the generator are spatially normalized at different levels of granularity, and $(ii)$ that the information on the targeted semantic layout and desired appearance of the synthesized output is propagated efficiently throughout the generator. %, and, as we show later, $(iii)$ is paramount for visually convincing virtual try-on results. 

%Each ResNet blocks in the generator stacks a sequence of two batch normalization and convolutional layers with the context-aware normalization placed  
Each ResNet block of the generator has two inputs: the image context $IC$, and the activation map from the previous model layer. The only exception here is the first ResNet block of the generator that uses image context $IC$ at the smallest resolution ($8 \times 6$ pixels) for both inputs, as shown in Figure~\ref{fig:cag}(a). The utilized ResNet blocks 
consist of a sequence of batch-normalization and convolutional layers repeated twice with CAN operations preceding the convolutional layers. If the output of the batch normalization is denoted as $X_{BN}$, then the context-aware normalization can formally be defined as:
%\begin{equation}
    $X_{CAN} = X_{BN} \odot \gamma + \beta,$
%\end{equation}
where $\odot$ denotes the Hadamard product and $X_{CAN}$ is the normalized output. %result of normalization operation. 
$\gamma$ and $\beta$ stand for (spatial) scale and bias parameters with the same dimensionality as $X_{BN}$. The parameters are learned during the training procedure and computed using three convolutional layers, one of which is shared to first project $IC$ onto a joint embedding space before estimating the values of $\gamma$ and $\beta$ with distinct convolutional operations.

\textbf{Training Objectives.} Two types of losses are designed to learn the parameters of C-VTON's generator, i.e.: \vspace{-0.5mm}
\begin{itemize}[leftmargin=*]
\setlength\itemsep{-0.01em}
    \item A \textbf{perceptual loss} ($\mathcal{L}_{per}$) that encourages the generator to produce a virtual try-on result as close as possible  to the reference input image $I$ in terms of semantics. Because the loss assumes the desired target appearance $I_C$ is known, the image context $IC$ is constructed with target clothing $C$ that matches the one in the input image $I$, i.e.: % In other words, when computing $\mathcal{L}_{per}$, we try to swap the initial clothing in $I$ with the same (matching) target clothing, which allows us to define a perceptual loss to drive the learning procedure, i.e.: % of the following form:
    \begin{equation} \label{eq:vggloss2}
        \mathcal{L}_{per} = \sum_i^n \tau_i \lVert \phi_i({I}'_C) - \phi_i(I) \rVert_1,
    \end{equation}
    where $\phi_i(\cdot)$ are feature maps produced by a pretrained VGG19~\cite{simonyan2014very} model before each of the $n=5$ max-pooling layer, $\tau_i$ is the $i$-th balancing weight, and ${I}'_C$ is the try-on result generated with the matching target clothing. 
    \item Three \textbf{adversarial losses} defined through three discriminators, each aimed at realistic generation of different aspects of the final try-on image, i.e.:\vspace{-0.5mm}
    \begin{itemize}[leftmargin=2.3mm]
    \setlength\itemsep{-0.01em}
        \item The \textbf{segmentation discriminator} ($D_{seg}$), inspired by \cite{schonfeld2020you}, aims to ensure realistic body-part generation by predicting (per-pixel) segmentation maps $S$ and their origin (real or fake). Given an input image ($I$ or $I_C$), $D_{seg}$ outputs a $w\times h\times (d+1)$ dimensional tensor, where the first $d$ channels contain segmented body parts and the ($d+1$)-st channel encodes whether a pixel is from a real or generated data distribution. $D_{seg}$ is trained by minimizing a ($d+1$)-class cross-entropy loss, i.e.: 
        \begin{equation}
            \begin{aligned}
            \label{eq:sd:loss}
                \mathcal{L}_{D_{seg}} = &-\mathbb{E}_{(I, S)} \left[ \sum_{k=1}^d \alpha_k \left(S_{k} \odot \log D_{seg}(I)_{k} \right) \right] \\
                & -\mathbb{E}_{I_{C}} \left[ \log D_{seg}(I_{C})_{d+1} \right],
            \end{aligned}
        \end{equation}
        where the first (segmentation-related) term applies to the (real) input images $I$ and penalizes the first $d$ output channels, and the second term penalizes the last remaining channel generated from the synthesized image $I_C = CAG(IC)$.
        $\alpha_k$ is a balancing weight calculated as the inverse frequency of the body-part in the given channels of the segmentation map $S$, i.e., 
%\begin{equation}
    $\alpha_k = hw/\lfloor S_k \rfloor$, where $\lfloor \cdot \rfloor$ is a cardinality operator. The corresponding adversarial loss ($\mathcal{L}_{seg}$) for the generator %, where $I_C$ is treated as a real image, 
    is finally defined as:
    \begin{equation}
        \mathcal{L}_{seg} = - \mathbb{E}_{(I_{C}, S)} \left[\sum_{k=1}^{d} \alpha_k \left(S_{k} \odot \log D_{seg}(I_{C})_{k} \right)\right].
    \end{equation}
     \item The \textbf{matching discriminator} ($D_{mth}$) aims to encourage the generator to synthesize output images with the desired target clothing by predicting whether the target garment $C$ corresponds to the clothing being worn in either $I$ or $I_C$. Formally, we train $D_{mth}$ by minimizing the following learning objective:
\begin{equation}
\begin{aligned}
     \mathcal{L}_{D_{mth}} = &-\mathbb{E}_{(I, C)} \left[ \log D_{mth}(I, C) \right] \\
    & -\mathbb{E}_{(I_{C}, C)} \left[ \log (1 - D_{mth}(I_{C}, C)) \right],
\end{aligned}
\end{equation}
leading to the following generator loss ($\mathcal{L}_{mth}$):
\begin{equation}
    \mathcal{L}_{mth} = - \mathbb{E}_{(I_{C}, C)} \left[ \log D_{mth}(I_{C}, C) \right].
\end{equation}
%Above, $I$ and $I_s$ correspond to original and generated images, while $C_t$ is in both cases an image of a clothing a person is wearing.
%\end{equation}
    \item The \textbf{patch discriminator} ($D_{ptc}$) contributes towards realistic body-part generation by focusing on the appearances of local patches, $P = \{p_0, ..., p_m\}, p_i \in \mathbb{R}^{w_p \times h_p \times 3}$, centered at $m=5$ characteristic body-parts, i.e., the neck, and both upper arms and forearms. Different from PatchGAN~\cite{isola2017image}, where patches are extracted implicitly through convolutional operations in the discriminator, we sample the patches from fixed locations based on the segmentation map $S$.
    The discriminator is trained to distinguish between real and generated body areas based on the following objective:
    \begin{equation}
    \begin{aligned}
         \mathcal{L}_{D_{ptc}} = &-\mathbb{E}_{P_{real}} \left[ \log D_{ptc}(P_{real}) \right] \\
        & -\mathbb{E}_{P_{fake}} \left[ \log (1 - D_{ptc}(P_{fake})) \right],
    \end{aligned}
    \label{eq:patchloss}
    \end{equation}
where $P_{real}$ and $P_{fake}$ correspond to patches extracted from the real and generated images $I$ and $I_{C}$, respectively. The generator loss then takes the following form:
\begin{equation}
    \mathcal{L}_{ptc} = - \mathbb{E}_{P_{fake}} \left[ \log D_{ptc}(P_{fake}) \right].
\end{equation}
    \end{itemize}
\end{itemize}
The \textbf{joint objective} for training C-VTON's context-aware generator is a weighted sum of all loss terms, i.e.:
\begin{equation}  
%\begin{aligned}
    \mathcal{L}_G  = \lambda_{per} \mathcal{L}_{per} + \lambda_{seg} \mathcal{L}_{seg} \\
                   + \lambda_{mth} \mathcal{L}_{mth} + \lambda_{ptc} \mathcal{L}_{ptc},
%\end{aligned}
\label{eq: overall_objective_CAG}
\end{equation}
where $\lambda_{per}$, $\lambda_{seg}$, $\lambda_{mth}$ and $\lambda_{ptc}$ denote hyperparameters that determine the relative importance of each loss term.

\section{Experiments}

In this section we now present experiments %the datasets used for the assessment of C-VTON, provide implementation details and report experimental results to highlight the merits of our virtual try solution. Specifically, results are presented 
that: $(i)$ compare C-VTON to competing models, $(ii)$ demonstrate the impact of key components on performance, and $(iii)$ explore the characteristics of the proposed model. %  architecure setup, and compare our results with CP-VTON~\cite{wang2018toward}, CP-VTON+~\cite{minar2020cp}, ACGPN~\cite{yang2020towards}, PF-AFN~\cite{ge2021parser} and S-WUTON~\cite{issenhuth2020do}. After describing used datasets and implementation details in Subsections~\ref{sec:experiments:datasets} and \ref{sec:experiments:implementation}, our method is compared with other approaches numerically (Subsection \ref{sec:experiments:quant} and qualitatively (Subsection~\ref{sec:experiments:qual}). Where available, we used generated images provided by the authors of the original papers or generated the images ourselves, given the code and/or trained models were available. In Subsection~\ref{sec:experiments:ablation}, we quantitatively and qualitatively evaluate contributions of each module to the final result.
\subsection{Datasets} \label{sec:experiments:datasets}

Following prior work~\cite{ge2021parser,issenhuth2020do,minar2020cp,yang2020towards}, two datasets are selected for the experiments, i.e., VITON~\cite{han2018viton} and MPV~\cite{dong2019towards}. %Details on the datasets are given below.

% \textbf{VITON}~\cite{han2018viton} and \textbf{VITON-HD} are similar datasets only differing in image resolution; $256 \times 192$ and $512 \times 368$ pixels, respectively. Among the two, VITON-HD is only recently becoming more popular due to being more difficult to train from. They contain $14,221$ training and $2032$ testing pairs of images (i.e., featuring subjects and target clothing). For the experiments, duplicate images that are present in both the training and test sets are filtered out, leaving us with $8586$ image pairs in the training set and $416$ image pairs in the test set. Once the duplicates are removed, the test set contains unique images not seen during training and %our and other models were not trained on, which 
% allows for a fair comparison between different approaches.

\textbf{VITON}~\cite{han2018viton} is a popular dataset for evaluating virtual try-on solutions and consists of $14,221$ training and $2032$ testing pairs of images (i.e., subjects and target clothing) with a resolution of $256 \times 192$ pixels. For the experiments,  duplicate images in the training and test sets are filtered out, leaving $8586$ image pairs in the training set and $416$ image pairs in the test set. After duplicate removal, the test set contains unique images not seen during training and %our and other models were not trained on, which 
allows for a fair comparison between different approaches.

\textbf{MPV}~\cite{dong2019towards} represents another virtual try-on dataset with $35,687$ person images ($256 \times 192$) wearing $13,524$ unique garments. Different from VITON, MPV exhibits a higher degree of appearance variability with larger differences in zoom level and view point. For the experiments, the images in MPV are %Person images are f
prefiltered to feature only (close to) frontal views in accordance with standard methodology, e.g.,~\cite{ge2021parser,issenhuth2020do}. The final train and test sets contain $17,400$ paired and $3662$ unpaired person and clothing images.
\begin{table}[t]
\begin{center}
\resizebox{\columnwidth}{!}{%
\renewcommand{\arraystretch}{1.1}
\begin{tabular}{lllrr}
\hline\hline
Data & \multicolumn{1}{l}{Model}   & Published    & FID$\downarrow$   & LPIPS$\downarrow$ ($\mu\pm\sigma$) \\ \hline 
\multirow{5}{*}{\begin{turn}{90}VITON\end{turn}} & CP-VTON~\cite{wang2018toward}  & ECCV $2018$    & $47.36$  & $0.303 \pm 0.043$ \\ 
& CP-VTON+~\cite{minar2020cp}   & CVPRW $2020$      & $41.37$  & $0.278 \pm 0.047$ \\ 
& ACGPN~\cite{yang2020towards}  & CVPR $2020$      & $37.94$  & $0.233 \pm 0.047$ \\ 
& PF-AFN~\cite{ge2021parser}    & CVPR $2021$      & $27.23$  & $0.237 \pm 0.049$ \\ 
& C-VTON                 &  This work   & $\mathbf{19.54}$ & $\mathbf{0.108 \pm 0.033}$ \\ \hline 
\multirow{3}{*}{\begin{turn}{90}MPV\end{turn}} & S-WUTON~\cite{issenhuth2020do}  & ECCV $2020$      & $8.188$  & $0.161 \pm 0.070$ \\ 
& PF-AFN$^\dagger$~\cite{ge2021parser}  &  CVPR $2021$          & $6.429$ & n/a     \\ 
& C-VTON & This work                       & $\mathbf{4.846}$ & $\mathbf{0.073 \pm 0.039}$\\ \hline\hline
\multicolumn{5}{l}{\small $^\dagger$ As reported in the original publication\vspace{-2mm}}
\end{tabular}
}
\end{center}
\caption{Quantitative comparison of C-VTON and competing state-of-the-art models in terms of FID and LPIPS scores - lower is better, as also indicated by the corresponding arrows. %DP-VTON is the top performer on both datasets.
\vspace{-2mm}}
\label{tab:experiments:quant:fid}
\end{table}

\subsection{Implementation Details} \label{sec:experiments:implementation}

C-VTON is implemented in Python using PyTorch. Most modules utilized in the processing pipeline build on ResNet-like blocks~\cite{he2016deep} that consist of two conv+ReLU layers %  a convolutional layer followed by a ReLU activation function repeated twice, 
and a trainable shortcut connection. Architectural details for the main C-VTON components are given below.%Spectral normalization is used on all linear and convolutional layer weights across the generator and all discriminators. 

\textbf{The Body-Part Geometric Matcher} (BPGM) consists of $2$ encoders, $E_1$ and $E_2$, %to produce features $F_{E_1, E_2}$ corresponding to body-part segmentations $S$ and target garment $C_t$. These encoders are 
with $5$ stacked convolutional layers, followed by a downsampling operation, a ReLU activation function and batch normalization. The feature regressor $R$ is implemented with $4$ convolutional layers, each followed by a ReLU activation and batch normalization layers. An $18$-dimensional linear output layer is  used to obtain the parameters ($\theta$) for thin-plate spline transformation. % are obtained through an $18$-dimensional linear output layer. %This output corresponds to a $3\times3$  Having $2$ values ($x$ and $y$ coordinates) for each control point, this results in $3\times3$-sized grid used for Thin-Plate Spline transformations.

\textbf{The Context-Aware Generator} (CAG) consists of ResNet blocks with context-aware normalization added before every convolutional layer. We use $6$ such blocks %for the implementation of in CAG, 
each followed by an ($2\times$) upsampling layer. Contextual inputs are resized to match each block's input resolution. An exponential moving average (EMA) is applied over generator weights with a decay value of $0.9999$, similarly to \cite{sushko2020you}. 

\textbf{Discriminators}. The matching discriminator $D_{mth}$ is implemented with two encoders (one for  $C$ and one for $I_C$)  consisting of $6$ ResNet blocks each. The output of the encoders is concatenated and fed to a linear layer that produces the final output. The patch discriminator $D_{ptc}$ comprises $4$ ResNet blocks arranged in an encoder architecture, with a fully-connected layer on top. The segmentation discriminator $D_{seg}$ has an UNet~\cite{ronneberger2015u} encoder-decoder architecture and consists of a total of $12$ ResNet blocks. %$\mathcal{D}_{match}$ and $\mathcal{D}_{patch}$ have $1$ and $\mathcal{D}_{seg}$ has $2$ additional Resnet blocks when trained on VITON-HD to account for higher image resolution.
\begin{table}[t]
\begin{center}
\resizebox{0.9\columnwidth}{!}{%
\renewcommand{\arraystretch}{1.1}
\begin{tabular}{lllr}
\hline\hline
Dataset                 & Model                             & Published     & vs. C-VTON   \\ \hline
\multirow{4}{*}{VITON}  & CP-VTON~\cite{wang2018toward}     & ECCV $2018$   & $0.766$      \\ 
                        & CP-VTON+~\cite{minar2020cp}       & CVPRW $2020$  & $0.756$      \\ 
                        & ACGPN~\cite{yang2020towards}      & CVPR $2020$   & $0.674$      \\ 
                        & PF-AFN~\cite{ge2021parser}        & CVPR $2021$   & $0.527$      \\ \hline
MPV                     & S-WUTON~\cite{issenhuth2020do}    & ECCV $2020$   & $0.607$      \\ \hline\hline
\end{tabular}\vspace{-1mm}
}
\end{center}
\caption{Results of the human perceptual study reported in terms of the frequency C-VTON generated results were preferred over others. The study was conducted with $100$ randomly selected images for each dataset and $70$ human participants.%Our method outperforms others in human evaluations. Higher percentage means more participants found the images generated from our model more realistic compared to other approaches.
\vspace{-1mm}}
\label{tab:experiments:quant:human}
\end{table}

\begin{figure*}[t]
\begin{minipage}[b]{0.99\textwidth}
\begin{minipage}[b]{0.01\textwidth}
 \
\end{minipage}
\begin{minipage}[b]{0.09\textwidth}
\footnotesize Original
\end{minipage}
\begin{minipage}[b]{0.065\textwidth}
\footnotesize Target
\end{minipage}
\begin{minipage}[b]{0.08\textwidth}
\footnotesize CP-VTON
\end{minipage}
\begin{minipage}[b]{0.095\textwidth}
\footnotesize CP-VTON+
\end{minipage}
\begin{minipage}[b]{0.083\textwidth}
\footnotesize ACGPN
\end{minipage}
\begin{minipage}[b]{0.093\textwidth}
\footnotesize PF-AFN
\end{minipage}
\begin{minipage}[b]{0.105\textwidth}
\footnotesize Ours
\end{minipage}
\begin{minipage}[b]{0.09\textwidth}
\footnotesize Original
\end{minipage}
\begin{minipage}[b]{0.068\textwidth}
\footnotesize Target
\end{minipage}
\begin{minipage}[b]{0.1\textwidth}
\footnotesize S-WUTON
\end{minipage}
\begin{minipage}[b]{0.05\textwidth}
\footnotesize Ours
\end{minipage}\vspace{-2.5mm}
\begin{center}
\includegraphics[width=\linewidth, trim = 0 44mm 0 4.7mm, clip]{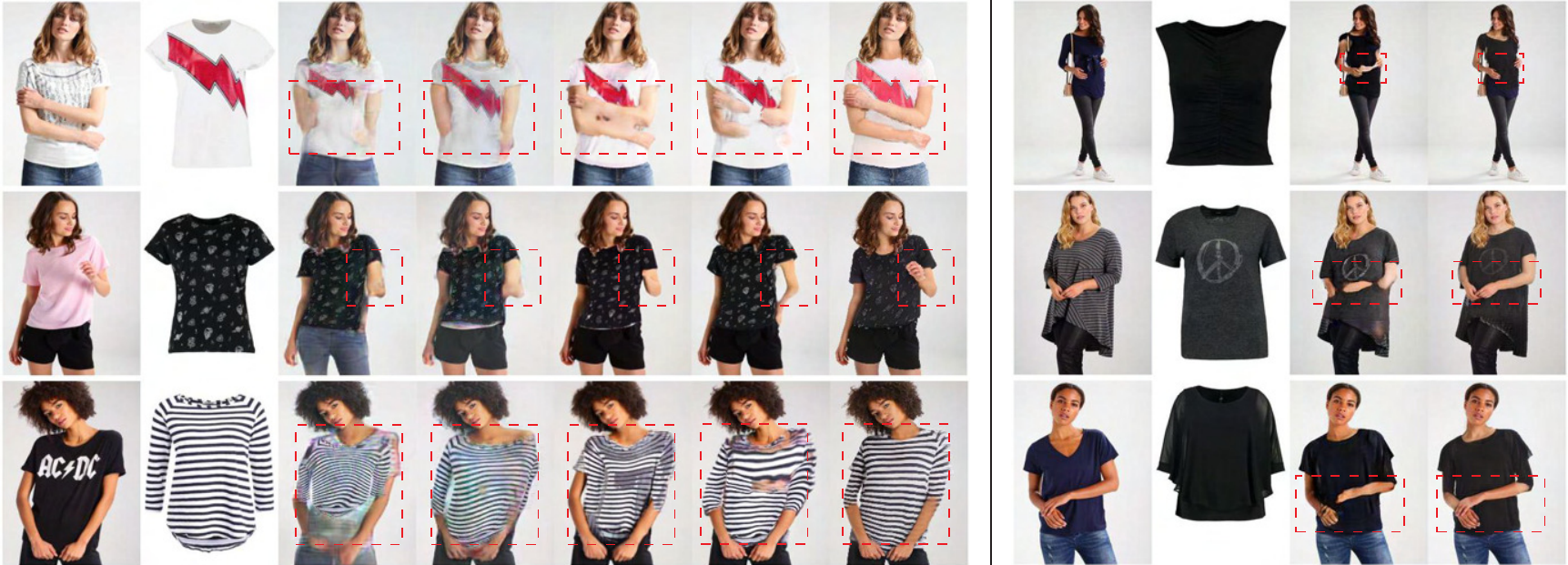}
\end{center}
\end{minipage}
   \caption{Comparison of C-VTON (ours) and several recent state-of-the-art models on the VITON (left) and MPV (right) datasets. Areas of interest in the synthesized images are marked with a red bounding box. C-VTON performs considerably better than competing models when synthesizing arms and hands and also better preserves on-shirt graphics. Best viewed electronically and zoomed-in for details.\vspace{-2mm}}
\label{fig:results:sota}
\end{figure*}

\textbf{Training Details.}  The ADAM optimizer~\cite{kingma2014adam} is used for the training procedure with a learning rate of $lr_{BPGM} = 0.0001$ for the BPGM, $lr_G = 0.0001$ for the generator and $lr_D = 0.0004$ for the discriminators. All weights in the learning objectives from Eqs.~\eqref{eq:vggloss}, \eqref{eq:bpgmjoint}, \eqref{eq:vggloss2} and \eqref{eq: overall_objective_CAG} are set to $1$, except for $\lambda_{vgg} = 0.1$ and $\lambda_{per} = 10$. The geometric matcher is trained for $30$ epochs and the generator for $100$ in all configurations. Source code is available at {\small\url{https://github.com/benquick123/C-VTON}}.
% Additional implementation details are available in the publicly released source code. % \footnote{{\scriptsize \url{https://github.com/benquick123/C-VTON}}.}

\subsection{Quantitative Results} \label{sec:experiments:quant}

To demonstrate the performance of C-VTON, we first analyze Fr\'echet Inception Distances (FID \cite{heusel2017gans}) and Learned Perceptual Image Patch Similarities (LPIPS)~\cite{zhang2018unreasonable} over processed VITON and MPV test images and conduct a human perceptual study (similarly to~\cite{ge2021parser, han2018viton, wang2018toward}) on the MTurk platform. For comparison purposes, we also report result for multiple state-of-the-art models, i.e., CP-VTON~\cite{wang2018toward}, CP-VTON+~\cite{minar2020cp}, ACGPN~\cite{yang2020towards}, PF-AFN~\cite{ge2021parser} and S-WUTON~\cite{issenhuth2020do}. Pretrained (publicly released) models are used for the experiments %in the experiments 
to ensure a fair comparison, except for S-WUTON, where synthesized test images were made available for scoring by the authors of the model. % Where available, we used generated images provided by the authors of the original papers or generated the images ourselves, given the code and/or trained models were available. In Subsection~\ref{sec:experiments:ablation}, we quantitatively and qualitatively evaluate contributions of each module to the final result.  

\textbf{FID and LPIPS Scores.} A quantitative comparison of C-VTON and the selected competitors is presented in %Our approach is evaluated with FID and LPIPS scores that can be seen in 
Table~\ref{tab:experiments:quant:fid}. We note that the results for PF-AFN on MPV are borrowed from~\cite{ge2021parser}, since no pretrained model is publicly available for this dataset. As can be seen, C-VTON significantly outperforms all competing models on both datasets. On VITON it reduces the FID score by $28.2\%$ compared to the runner-up and the LPIPS measure by $53.6\%$. Similar (relative) performances are also observed on MPV, where C-VTON again leads to comparable reductions in FID and LPIPS scores when compared to the runner-ups. We attribute these results to the simplified geometric matching procedure used in C-VTON and the inclusion of diverse contextual information in the final image synthesis step. % These results clearly demonstrate the benefits of using body part segmentation     %We show that our method outperforms others on both datasets, indicated by lower scores in all scenarios. In each scenario we report scores with relative differences 7.69 (FID) and 0.129 (LPIPS) on VITON, and 1.512 (FID) on MPV datasets, compared to PF-AFN, the second-best approach. Where the original code and trained models or generated test images are available, the scores are calculated using the same configuration to ensure fair comparisons between methods. Note, that FID scores on VITON dataset are higher compared to scores reported by the original authors due to modified number of samples in the test set as described in section~\ref{sec:experiments:datasets}.

\textbf{Human Perceptual Study}. We also evaluate C-VTON through a human perceptual study to analyze the (subjectively) perceived quality of the generated try-on images. In the scope of the study, participants were shown the original input image, the target garment and two distinct try-on results, where one was always the result of C-VTON and the other was generated by one of the competing solutions. The participants had to choose the more convincing of the two images based on multiple factors, i.e.: texture transfer quality, arm generation capabilities, pose preservation, and overall quality of results. $100$ randomly selected images from each dataset were used for the study, which featured $70$ participants in total. The results in Table~\ref{tab:experiments:quant:human}, reported in terms of frequency C-VTON generated results were preferred over others, show that the proposed approach was clearly favored among the human raters.% with the PF-AFN approach producing the closest   As can be seen from Table~\ref{tab:experiments:quant:human}, where the percentage of trials where images generated with C-VTON were preferred to others are reported, the proposed approach  . 
\begin{table}[t]
\begin{center}
\resizebox{\columnwidth}{!}{%
\renewcommand{\arraystretch}{1.15}
\begin{tabular}{lcclcc}
\hline\hline
\multirow{2}{*}{Model}             & \multicolumn{2}{c}{VITON} && \multicolumn{2}{c}{MPV} \\ \cline{2-3} \cline{5-6}
             & FID$\downarrow$    & LPIPS$\downarrow$     &  & FID$\downarrow$         & LPIPS$\downarrow$        \\ \hline \hline
C-VTON             & $\mathbf{19.535}$ & $\mathbf{0.108  \pm 0.033}$
                    && $\mathbf{4.846}$  & $\mathbf{0.073  \pm 0.039}$\\ \hdashline
A1: w/o CAN             & $24.521$ & $0.162  \pm 0.037$
                   & & $12.096$ & $0.159  \pm 0.049$ \\ 
A2: w/o BPGM            & $24.422$ & $0.140  \pm 0.036$
                    && $6.728$  & $0.096  \pm 0.046$ \\ 
A3: w/o $\mathcal{D} ^\dagger$   & $21.359$ & $0.109  \pm 0.033$
                    && $5.898$  & $0.076  \pm 0.040$ \\ 
A4: w/o EMA             & $24.571$ & $0.150  \pm 0.035$
                    && $5.304$  & $0.102 \pm 0.043$ \\ 
 \hline\hline
 \multicolumn{6}{l}{$^\dagger$ \small $\mathcal{D}$ stands for the set of discriminators $\mathcal{D} =\{D_{seg}, D_{mth}, D_{ptc}\}$\vspace{-2mm}}
\end{tabular}\vspace{-4mm}
}
\end{center}
\caption{Ablation study results. For each C-VTON variant (A1-A4) one key component is ablated to demonstrate its contribution. %Lower scores imply better performance.%DP-VTON achieves the lowest FID and LPIPS scores, suggesting that all components are critical for the final results. = \{D_{seg}, D_{match}, D_{patch}\}
\vspace{-4mm} }
\label{tab:experiments:ablation}
\end{table}
\begin{figure}[t]
\begin{minipage}[b]{0.475\textwidth}
\begin{minipage}[b]{0.01\textwidth}
\scriptsize \
\end{minipage}
\begin{minipage}[b]{0.145\textwidth}
\scriptsize Original
\end{minipage}
\begin{minipage}[b]{0.12\textwidth}
\scriptsize Target
\end{minipage}
\begin{minipage}[b]{0.117\textwidth}
\scriptsize w/o CAN
\end{minipage}
\begin{minipage}[b]{0.165\textwidth}
\scriptsize w/o BPGM
\end{minipage}
\begin{minipage}[b]{0.105\textwidth}
\scriptsize w/o $\mathcal{D}$
\end{minipage}
\begin{minipage}[b]{0.130\textwidth}
\scriptsize w/o EMA
\end{minipage}
\begin{minipage}[b]{0.13\textwidth}
\scriptsize C-VTON
\end{minipage}\vspace{-2mm}
\begin{center}
\includegraphics[width=\linewidth, trim = 0 0 0 8mm,clip]{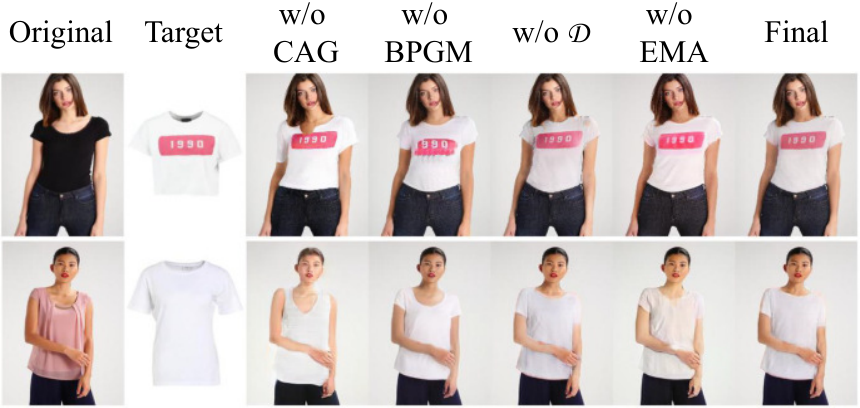}
\end{center}
\end{minipage}
%\hfill
%\begin{minipage}[b]{0.493\textwidth}
%\begin{center}
%\includegraphics[width=\linewidth, trim = 88mm 0 0 0,clip]{fig/results_ablation.pdf}
%\end{center}
%\end{minipage}
   \caption{Qualitative ablation-study results. %comparing generated images when omitting different parts of our architecture on VITON (left) and MPV (right) datasets. 
   %The removal of any component leads to less convincing synthesis results.
   The aggregation of all component results in  
   %The most 
   noticeable improvements in sleeve and arm generation, on-garment graphics  and realistic garment shapes.\vspace{-4mm}}
\label{fig:results:ablation}
\end{figure}

\subsection{Qualitative Results} \label{sec:experiments:qual}

% \begin{figure}[t]
% \begin{minipage}[b]{0.9\columnwidth}
% \begin{minipage}[b]{0.02\columnwidth}
% \scriptsize \
% \end{minipage}
% \begin{minipage}[b]{0.165\columnwidth}
% \scriptsize Original
% \end{minipage}
% \begin{minipage}[b]{0.175\columnwidth}
% \scriptsize Target
% \end{minipage}
% \begin{minipage}[b]{0.122\columnwidth}
% \scriptsize CS
% \end{minipage}
% \begin{minipage}[b]{0.175\columnwidth}
% \scriptsize S-w-CS
% \end{minipage}
% \begin{minipage}[b]{0.125\columnwidth}
% \scriptsize BPS
% \end{minipage}
% \begin{minipage}[b]{0.135\columnwidth}
% \scriptsize S-w-BPS
% \end{minipage}\vspace{-2.5mm}
% \begin{center}
% \includegraphics[width=\linewidth]{fig/results_clothseg_additional.pdf}
% \end{center}
% \end{minipage}
%    \caption{Impact of segmentation procedure on synthesis results. CS: clothing segmentation, S-w-CS: Synthesis with CS, BPS - Body-part segmentation, S-w-BPS: Synthesis with BPS. % Comparisons of effects of two types of segmentations on VITON dataset. Use of clothing segmentations affects the sleeve generation, garment edges and arm generation in difficult poses.
%    }\vspace{-2.5mm}
% \label{fig:viton:seg}
% \end{figure}

Next, we explore the performance of C-VTON through visual comparisons with competing models in Figure~\ref{fig:results:sota}. % Additionally to numerical results, our generated images are also visually compared with virtual try-on from other approaches. Figure~\ref{fig:results:sota} shows comparison with the same set of methods as in the previous section. 
Due to the unavailability of a pretrained PF-AFN model for MPV, C-VTON is only compared to S-WUTON on this dataset. As can be seen from the presented examples, the proposed approach generates the most convincing virtual try-on results and %fares . Generally, we at least match results from other approaches, but as can be seen on the Figure~\ref{fig:results:sota} our model 
performs particularly well with hand and on-shirt graphics synthesis. The results clearly show that %photo realistic and 
visually convincing virtual try-on results can be produced with C-VTON even with subjects imaged in difficult poses and with challenging arm/hand configurations.

Among the evaluated competitors, PF-AFN produces the most convincing results on the VITON dataset. However, as illustrated by the presented examples, 
%In general, the qualitative examples corresponds well to the quantitative results discussed above in terms of relative ranking of the evaluated methods. PF-AFN results, the second-best approach on VITON dataset, are in line with reported quantitative results. In practice 
the method sometimes does not preserve arms, the initial body shape and/or the pose of the subjects, whereas C-VTON fares much better in this regard. The remaining approaches, i.e., CP-VTON, CP-VTON+ and ACGPN, produce less convincing results and often fail to preserve certain (non-transferable) image parts (e.g. trousers and skirts) and textures from the target garment. On MPV, S-WUTON similarly struggles to preserve arms and body shape, while our model synthesizes both well. The excellent performance of C-VTON in this regard is the results of the body-part segmentation procedure used and the set of carefully designed discriminators that ensure realism of the generated images. %a and not  due to the reliance on body part segmentatations . More generated images and comparisons are available in Appendices~\ref{appendix:generated} and \ref{appendix:comparisons}. Images presenting model limitations are available in Appendix~\ref{appendix:limitations}.
%Results on VITON-HD dataset are available in the supplementary material.

%Contributions of our BPGM compared to regular GMM and usage of clothing versus DensePose segmentations are visible in Figure~\ref{fig:effect}, where it can be seen that the warping quality and clothing-agnostic segmentations critically influence the end result. More examples can be found in Appendix~\ref{appendix:effect}.
\begin{figure}[t]
\begin{center}
\includegraphics[width=0.9\columnwidth]{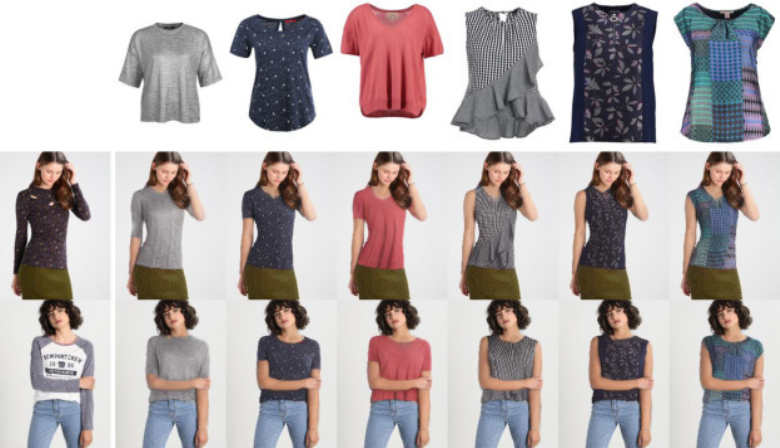}
\end{center}\vspace{-2.5mm}
   \caption{Sample results with multiple target garments and different subjects. Challenging examples with long to short-sleeved garment transfer are presented. Best viewed zoomed-in for details.}\vspace{0mm}
\label{fig:long2short}
\end{figure}
\begin{figure}[t]
\begin{minipage}[b]{0.9\columnwidth}
\begin{minipage}[b]{0.02\columnwidth}
\scriptsize \
\end{minipage}
\begin{minipage}[b]{0.165\columnwidth}
\scriptsize Original
\end{minipage}
\begin{minipage}[b]{0.125\columnwidth}
\scriptsize Target
\end{minipage}
\begin{minipage}[b]{0.152\columnwidth}
\scriptsize GMM out
\end{minipage}
\begin{minipage}[b]{0.155\columnwidth}
\scriptsize S-w-GMM
\end{minipage}
\begin{minipage}[b]{0.150\columnwidth}
\scriptsize BPGM out
\end{minipage}
\begin{minipage}[b]{0.165\columnwidth}
\scriptsize S-w-BPGM
\end{minipage}\vspace{-2.5mm}
\begin{center}
\includegraphics[width=\linewidth]{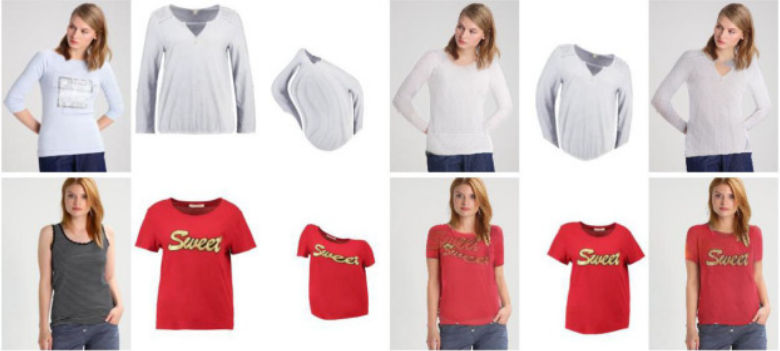}
\end{center}
\end{minipage}
   \caption{Comparison of the geometric matching module (GMM) from \cite{wang2018toward} and our Body-Part Geometric matcher (BPGM). S-w-GMM: Synthesis with GMM, S-w-BPS: Synthesis with BPGM.
   % CS: clothing segmentation, S-w-CS: Synthesis with CS, BPS - Body-part segmentation, S-w-BPS: Synthesis with BPS. % Comparisons of effects of two types of segmentations on VITON dataset. Use of clothing segmentations affects the sleeve generation, garment edges and arm generation in difficult poses.
   }\vspace{-2.5mm}
\label{fig:viton:warp}
\end{figure}

\subsection{Ablation Study}  \label{sec:experiments:ablation}

C-VTON relies on several key components to facilitate image-based virtual try-on. 
To demonstrate the impact of these components on performance, an ablation study is conducted. Specifically, four C-VTON variants and implemented, %trained and evaluated, %each without one of the key components, 
i.e.: $(i)$ C-VTON without CAN operations (A1), $(ii)$ C-VTON without the BPGM (A2), $(iii)$ C-VTON without the discriminators (A3), and $(iv)$ C-VTON without the exponential moving average - EMA (A4). 

The results in Table~\ref{tab:experiments:ablation} show that FID and LPIPS scores increase when any of the %considered 
key components is ablated. %This suggests that all part of the model are critical for the overall performance even though with different contributions.  
%and Figure~\ref{fig:results:ablation} show
%In this subsection, we show qualitative and quantitative improvement corresponding to usage of different DP-VTON modules. Experiments with removed modules listed in Table~\ref{tab:experiments:ablation} were performed to evaluate their contribution to the final result. It can be seen that FID and LPIPS scores increase when any of the specified modules is absent.
Interestingly, the absence of CAN operations seems to affect results the most, while the discriminators appear to contribute the least. However, when looking at the visual examples in Figure~\ref{fig:results:ablation}, we see that the discriminators critically affect the final image quality despite the somewhat smaller change in quantitative scores. The difference is especially noticeable when comparing sharpness and artefacts when training C-VTON with or without the discriminators. Furthermore, CAN operations contribute to sleeve and arm generation, the BPGM to a higher quality of on-garment graphics, and EMA to more realistic garment shapes. C-VTON combines these contributions without creating new artefacts, as illustrated by the results for the complete model.
\begin{figure}[t]
\begin{minipage}[b]{1\columnwidth}
\begin{minipage}[b]{0.008\textwidth}
\tiny \
\end{minipage}
\begin{minipage}[b]{0.07\columnwidth}
\scriptsize Orig.
\end{minipage}
\begin{minipage}[b]{0.09\columnwidth}
\scriptsize Target
\end{minipage}
\begin{minipage}[b]{0.075\columnwidth}
\scriptsize \cite{wang2018toward}
\end{minipage}
\begin{minipage}[b]{0.08\columnwidth}
\scriptsize \cite{minar2020cp}
\end{minipage}
\begin{minipage}[b]{0.09\columnwidth}
\scriptsize \cite{yang2020towards}
\end{minipage}
\begin{minipage}[b]{0.061\columnwidth}
\scriptsize \cite{ge2021parser}
\end{minipage}
\begin{minipage}[b]{0.105\columnwidth}
\scriptsize Ours
\end{minipage}
\begin{minipage}[b]{0.075\columnwidth}
\scriptsize Orig.
\end{minipage}
\begin{minipage}[b]{0.095\columnwidth}
\scriptsize Target
\end{minipage}
\begin{minipage}[b]{0.065\columnwidth}
\scriptsize \cite{issenhuth2020do}
\end{minipage}
\begin{minipage}[b]{0.07\columnwidth}
\scriptsize Ours
\end{minipage}
\vspace{-2mm}
\begin{center}
\includegraphics[width=0.98\columnwidth,trim = 2.2mm 0 2.2mm 5mm, clip]{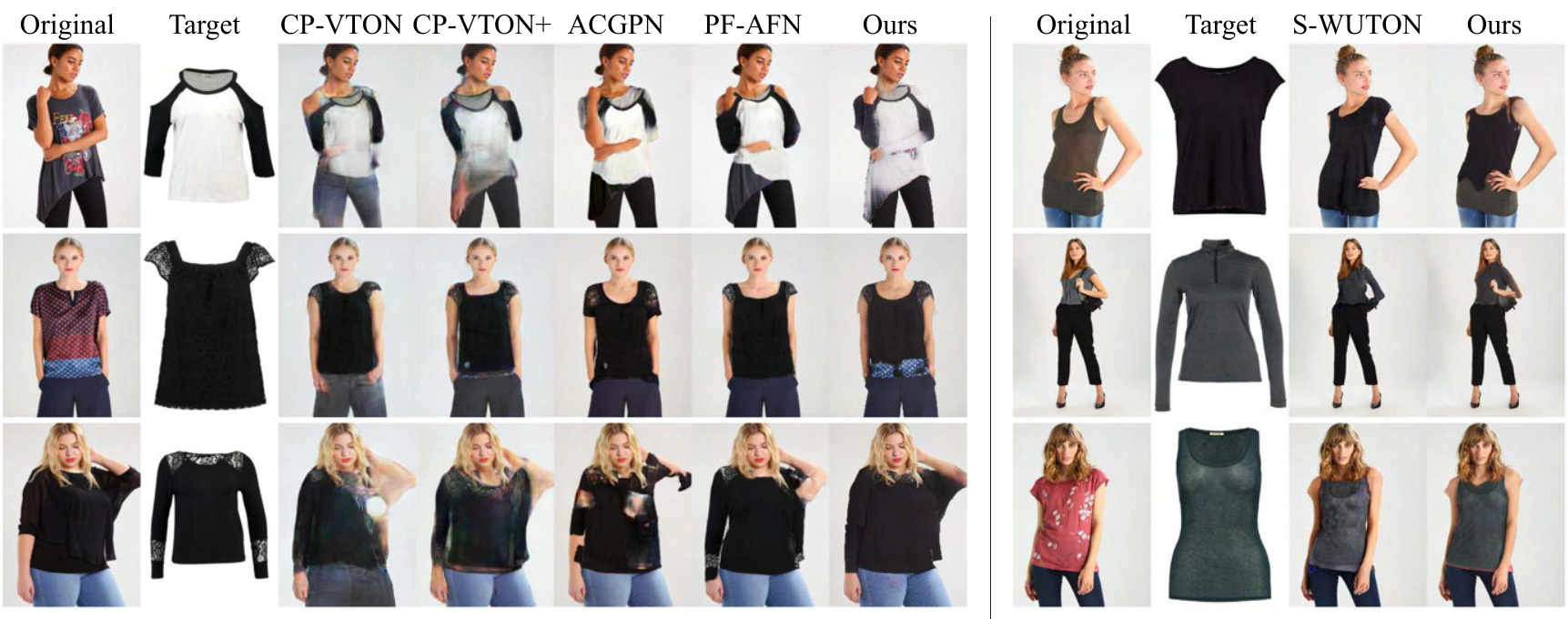}
\end{center}
\end{minipage}
   \caption{Examples of less convincing try-on results. With certain image characteristics C-VTON generates blurry clothing edges and only partially transfers target garments. Zoom in for details. %  transferred. Note, however, that other methods have problems with realistic synthesis of above images as well.
   }
\label{fig:limit}\vspace{-1mm}
\end{figure}
\subsection{Strengths and Weaknesses}  \label{sec:experiments:saw}

Finaly, we demonstrate some of  C-VTON's strengths and weaknesses by: $(i)$ presenting virtual try-on results for multiple target garments and different subjects, $(ii)$  highlighting the benefits of the proposed geometric matcher, and $(iii)$ illustrating some of the model's limitations.

\textbf{Multiple Targets and Subjects.} Figure~\ref{fig:long2short} shows visual try-on results for two distinct subjects and $6$ different target garments with varying sleeve lengths. Note that despite the fact that the arms are completely covered in the input images, C-VTON is able to generate realistic try-on results with convincing arm appearances. Varying sleeve lengths are also transferred well onto the synthesized images.  

% \textbf{Body-part vs. clothing segmentation.} Figure~\ref{fig:viton:seg} illustrates the benefit of conditioning C-VTON on body-part (BPS) rather than clothing (CS) segmentations. As can be seen, clothing segmentations often result in incorrect shapes of the target garments in the generated images images (see S-w-CS), where only the texture is transferred correctly. 
%The results in the top row show how errors in clothing segmentation affect the synthesis step. Here, the synthesis step produces visual artefacts due to the segmentation errors (S-w-CS), which are not present in body-part conditioned and the results in bottom-row show how demonstrate that even with correct clothing segmentations the shape of the initial clothing item is preserved and only the texture is transfered to the output (see S-w-CS results). 
% Conditioning C-VTON on body-part segmentations (S-w-BPS) helps generating correct target-clothing shapes and leads to better synthesis of initially occluded image areas.    

\textbf{GMM vs. BPGM.} Figure~\ref{fig:viton:warp} shows a comparison between the original geometric matching module (GMM) from \cite{wang2018toward} and the proposed body-part geometric matcher (BPGM). Note that our BPGM generates more realistic warps that better preserve the shape and texture of the target clothing in the final try-on result. As illustrated by the example in the top row, the better alignment ensured by the BPGM leads to a correctly rendered V-neck. Similarly, on-shirt graphics are better preserved when the proposed BPGM is used instead of the original GMM, as seen by the example in the bottom row of Figure~\ref{fig:viton:warp}. % and more convincing on-shirt graphics. In the second case, the provided example showcases that incorrect graphic warping persists in the final image and even gets replaced by texture on the non-warped target clothing when using GMM, whereas our BPGM doesn't have this effect on the final image.

\textbf{Limitations.}  Issues with the masking procedure (of $I_m$) when generating the image context $IC$, loose clothing in the input images, and the inability of the model to differentiate between the front and backside of the target garment $C$ are among the main causes for some of the less convincing virtual try-on results produced with C-VTON. These causes lead  to  unrealistic  and  soft  garment  edges, incorrectly synthesized clothing types and improperly rendered neck areas. Similar limitations are also observed with the competing models, as seen from the presented examples in Figure~\ref{fig:limit}. 

%------------------------------------------------------------------------
\section{Conclusion}

%While noisy clothing segmentations were known to be a bottleneck in virtual try-on image generation, we show that body-part segmentation maps, utilized at different points in our model improve the final image quality. 
In this paper, we proposed C-VTON, a novel approach to image-based virtual try-on capable of synthesizing high-quality try-on results across a wide range of input-image characteristics. The model was evaluated in extensive experiments on the VITON and MPV datasets and was shown to clearly outperform the state-of-the-art. Additional results that further highlight the merits of the proposed approach are available in the Supplementary Material.

%Driven by that insight, we propose Context-Driven Virtual Try-On Network (C-VTON) that synthesizes high-quality images. In addition to efficient use of segmentation maps, we present Body-Part Geometric Matcher (BPGM) that warps clothing without producing unwanted artefacts, Context-Aware Normalization for efficient use of auxiliary information and use multiple discriminator architectures to generate realistic results and improve sharpness. Our model outperforms other state-of-the-art models in this field especially at arm generation in challenging poses and oveimprs preservation of on-garment textures.

%\clearpage

\section*{Acknowledgements}\label{Sec:ack}

Supported in parts by the Slovenian Research Agency ARRS through the Research Programmes P2-0250(B) "Metrology and Biometric System" and P2--0214 (A) “Computer Vision”, the ARRS Project J2-2501(A) "DeepBeauty" and the ARRS junior researcher program.

{\small
\bibliographystyle{ieee_fullname}
\bibliography{egbib}
}

\clearpage

\appendix

\begin{Large}
   \noindent \textbf{C-VTON Supplementary Material}
\end{Large}
\\ %\vspace{5mm}

\begin{large}
 %  \noindent \textbf{Paper ID: 1065}
\end{large}
%\\ %\vspace{5mm}

\begin{figure*}[t]
\begin{center}
\includegraphics[width=0.97\linewidth]{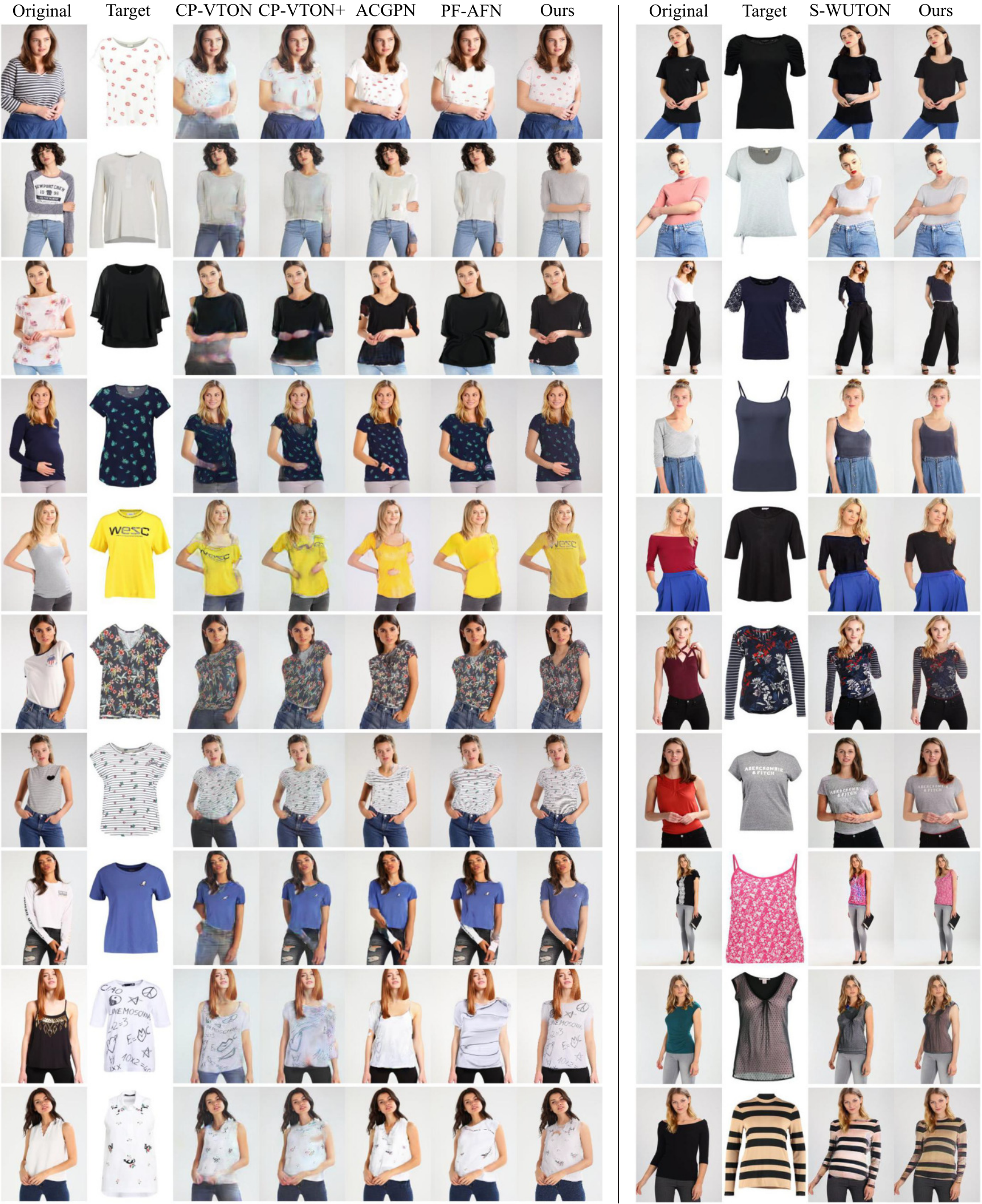}
\end{center}
   \caption{Additional comparisons on the VITON (left) and MPV (right) datasets. The proposed C-VTON model fares exceedingly well when synthesizing people in difficult positions and with detailed on on-garment graphics. The figure is best viewed electronically and zoomed-in for details.%In other cases, we tend to match the performance of the second-best approaches.
   }
\label{fig:appendix:comparisons}
\end{figure*}

In the main part of the paper we presented a wide variety of results to highlight the benefits of the proposed C-VTON virtual try-on approach. In this \textit{Supplementary material} we now show additional visual result to further demonstrate the capabilities of C-VTON. Specifically, we: $(i)$ show comparisons with competing models from the literature using a wider range of input images, $(ii)$ present additional try-on results in different settings, e.g., under pose variations, with varying sleeve lengths, or with differently textured garments, $(iii)$ investigate the performance of C-VTON on an additional high-resolution dataset, i.e., VITON-HD, $(iv)$ study the impact of body-part segmentations on the overall performance, $(v)$ compare results between the Geometric Matching Module from ~\cite{wang2018toward} and our Body-Part Geometric Matcher, and $(vi)$ provide a more in-depth analysis of the limitations of the proposed (virtual try-on) model.     

\section{Additional Comparisons} \label{appendix:comparisons}

Figure~\ref{fig:appendix:comparisons} shows additional comparisons between C-VTON and competing approaches from the literature. Here, the same models as in the main part of the paper are again used for the comparison. As can be seen, C-VITON consistently outperforms others in arm and texture generation, both on the VITON as well as on the MPV dataset. Additionally, it can be seen that on MPV, S-WUTON seems to produce images with extreme color saturation in certain cases, e.g. see results in rows $8$ and $10$. C-VTON does not suffer from such issues.

\section{Additional Try-On Examples} \label{appendix:generated}

\begin{figure*}[t]
\begin{center}
\includegraphics[width=0.84\linewidth]{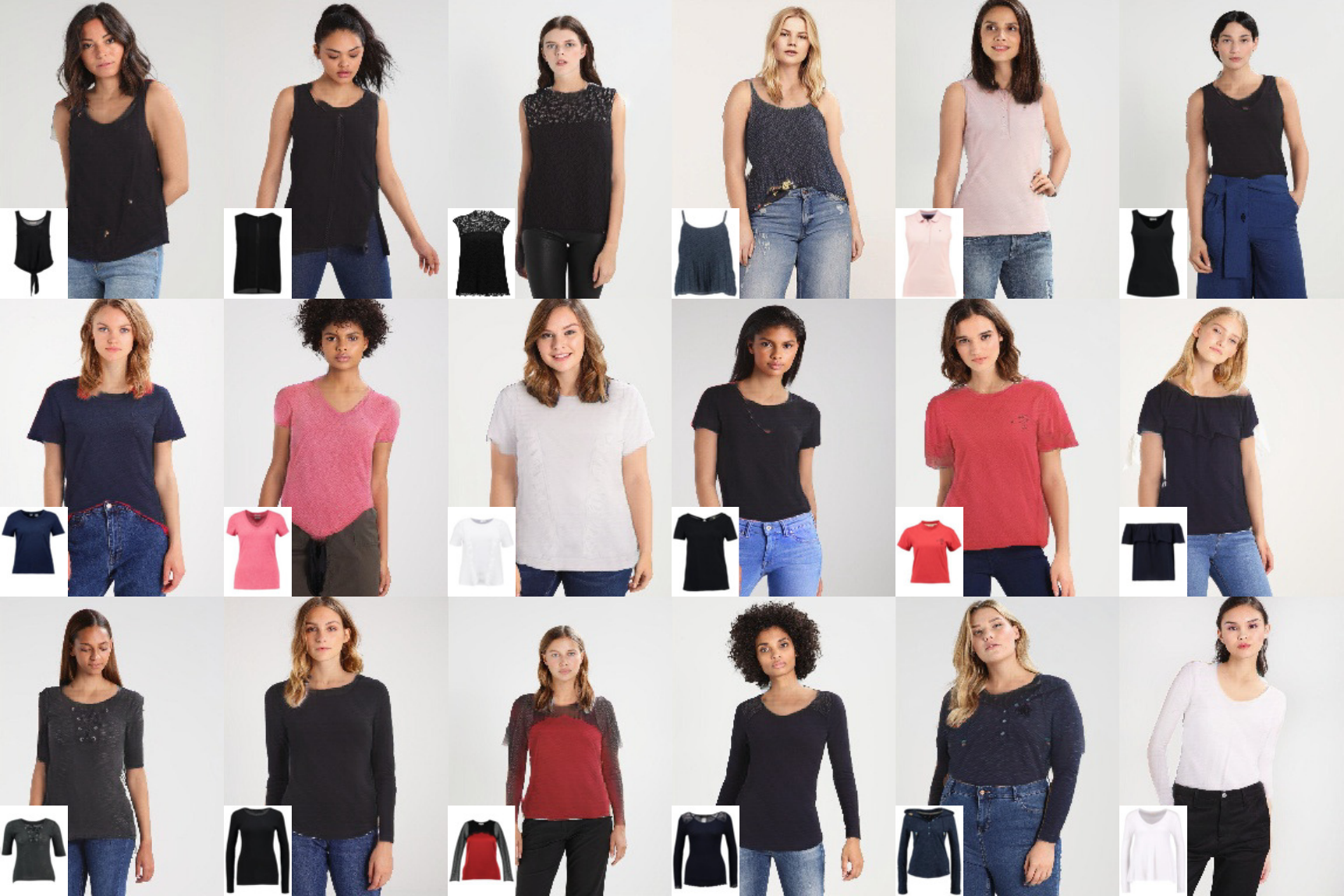}
\end{center}\vspace{-2mm}
   \caption{Example results from the VITON test dataset containing simple garments and people in simple poses ordered by sleeve length. The image in the bottom left of each example shows the target garment. The figure is best viewed in color.}
\label{fig:appendix:viton:basic}%\vspace{2mm}
\end{figure*}

\begin{figure*}[t]
\begin{center}
\includegraphics[width=0.84\linewidth]{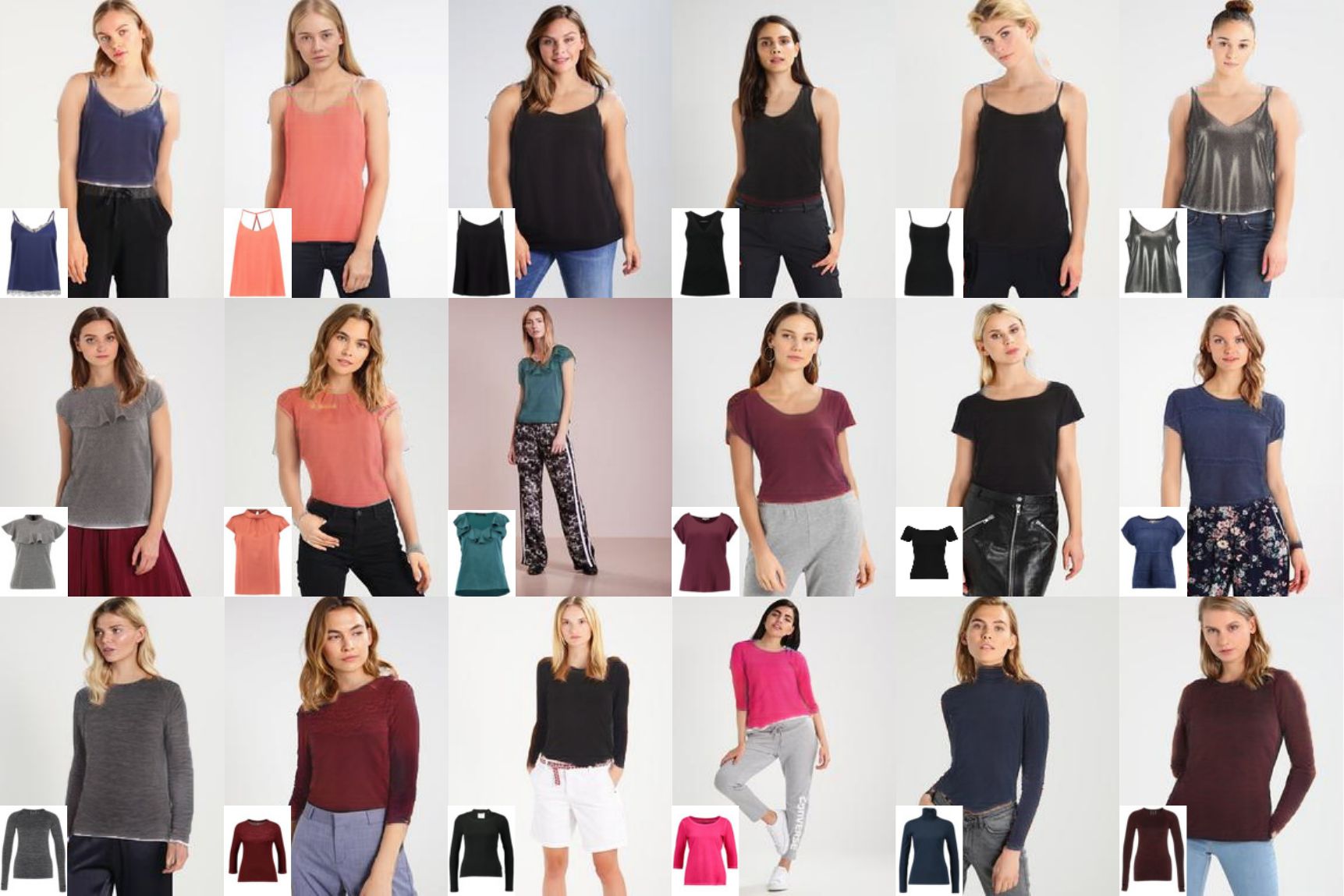}
\end{center}\vspace{-2mm}
   \caption{Example results from the MPV test dataset containing simple garments and people in simple poses ordered by sleeve length. The image in the bottom left of each example shows the target garment. The figure is best viewed in color.}
\label{fig:appendix:mpv:basic}
\end{figure*}

\begin{figure*}[t]
\begin{center}
\includegraphics[width=0.92\linewidth]{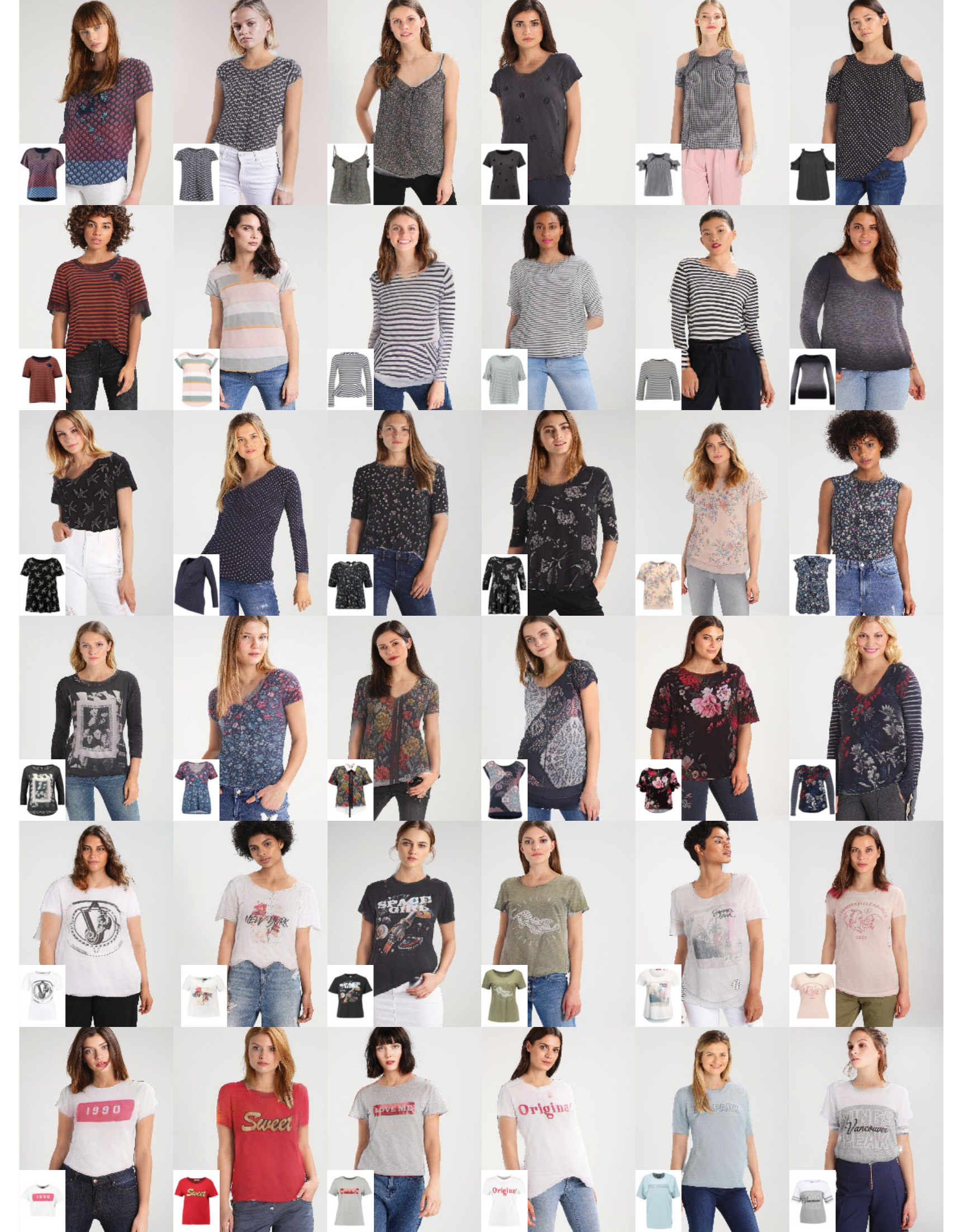}
\end{center}
   \caption{Example results from the VITON test dataset containing textured garments and people in simple poses. The type of on-garment texture displayed differs from row to row to show our model's ability to synthesize a wide array of graphics. The image in the bottom left of each example shows the target garment. Zoom in for details.}
\label{fig:appendix:viton:texture}
\end{figure*}

\begin{figure*}[t]
\begin{center}
\includegraphics[width=0.92\linewidth]{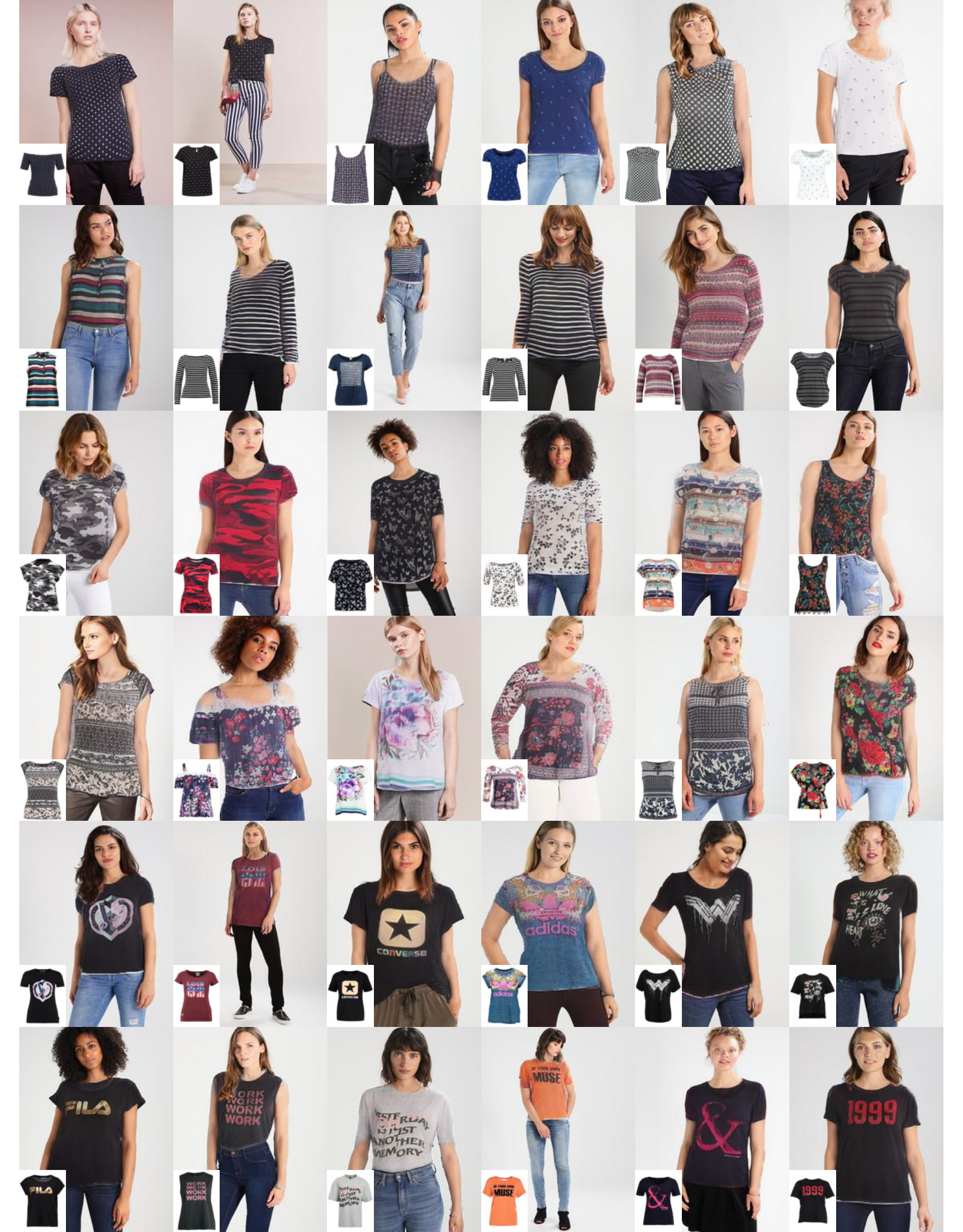}
\end{center}
   \caption{Examples results from the MPV test dataset containing textured garments and people in simple poses. The type of on-garment texture displayed differs from row to row to show our model's ability to synthesize wide array of graphics. The image in the bottom left of each example shows the target garment. Zoom in for details.}
\label{fig:appendix:mpv:texture}
\end{figure*}

\begin{figure*}[t]
\begin{center}
\includegraphics[width=\linewidth]{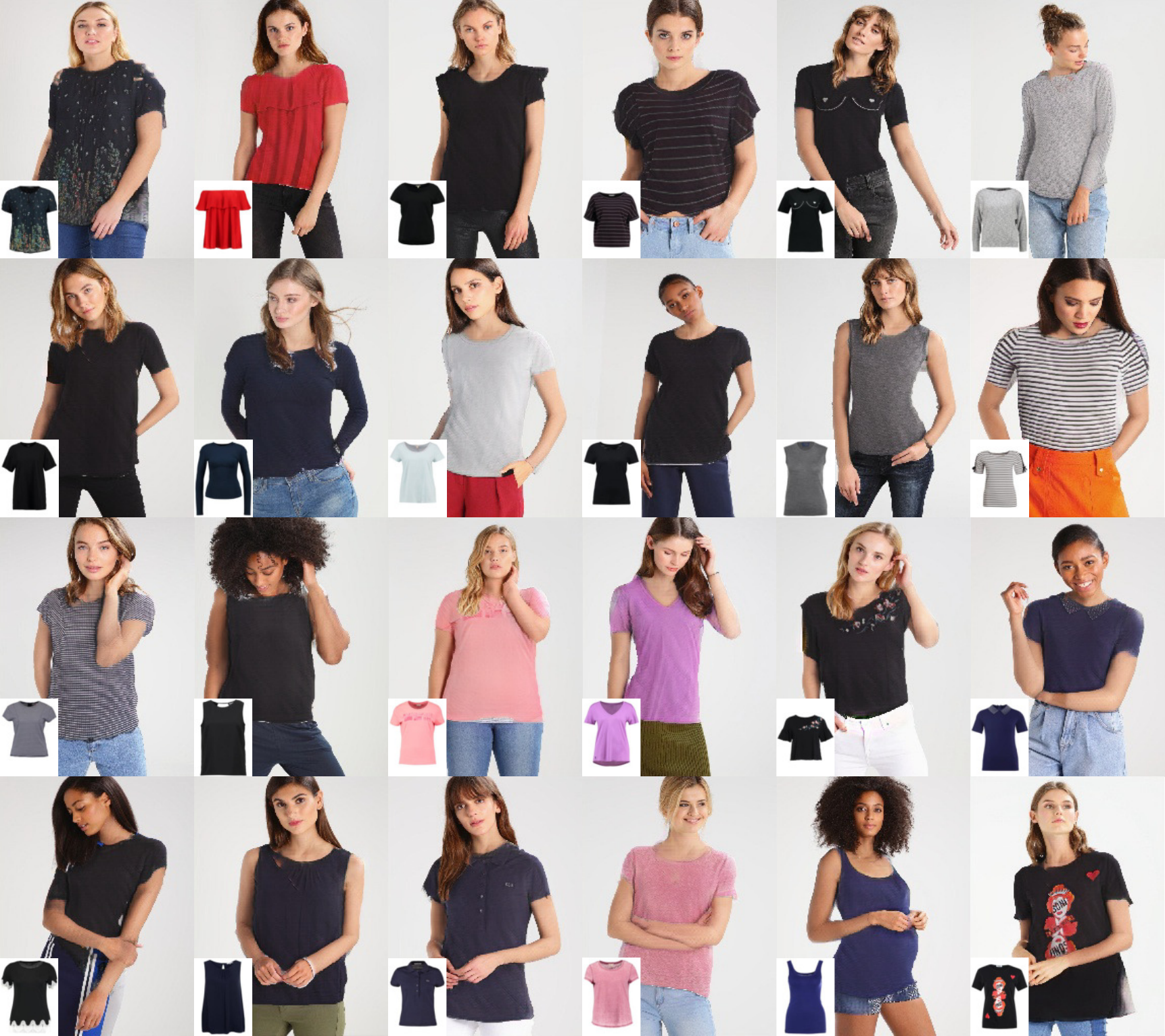}
\end{center}
   \caption{Examples results from the VITON test set containing people in what we consider difficult poses. The generated images are arranged into $4$ distinct poses to show the consistency of the synthesis procedure in a wide variety of conditions. The image in the bottom left of each example, shows the target garment. The figure is best viewed in color and electronically. }
\label{fig:appendix:viton:pose}
\end{figure*}

\begin{figure*}[t]
\begin{center}
\includegraphics[width=0.99\linewidth]{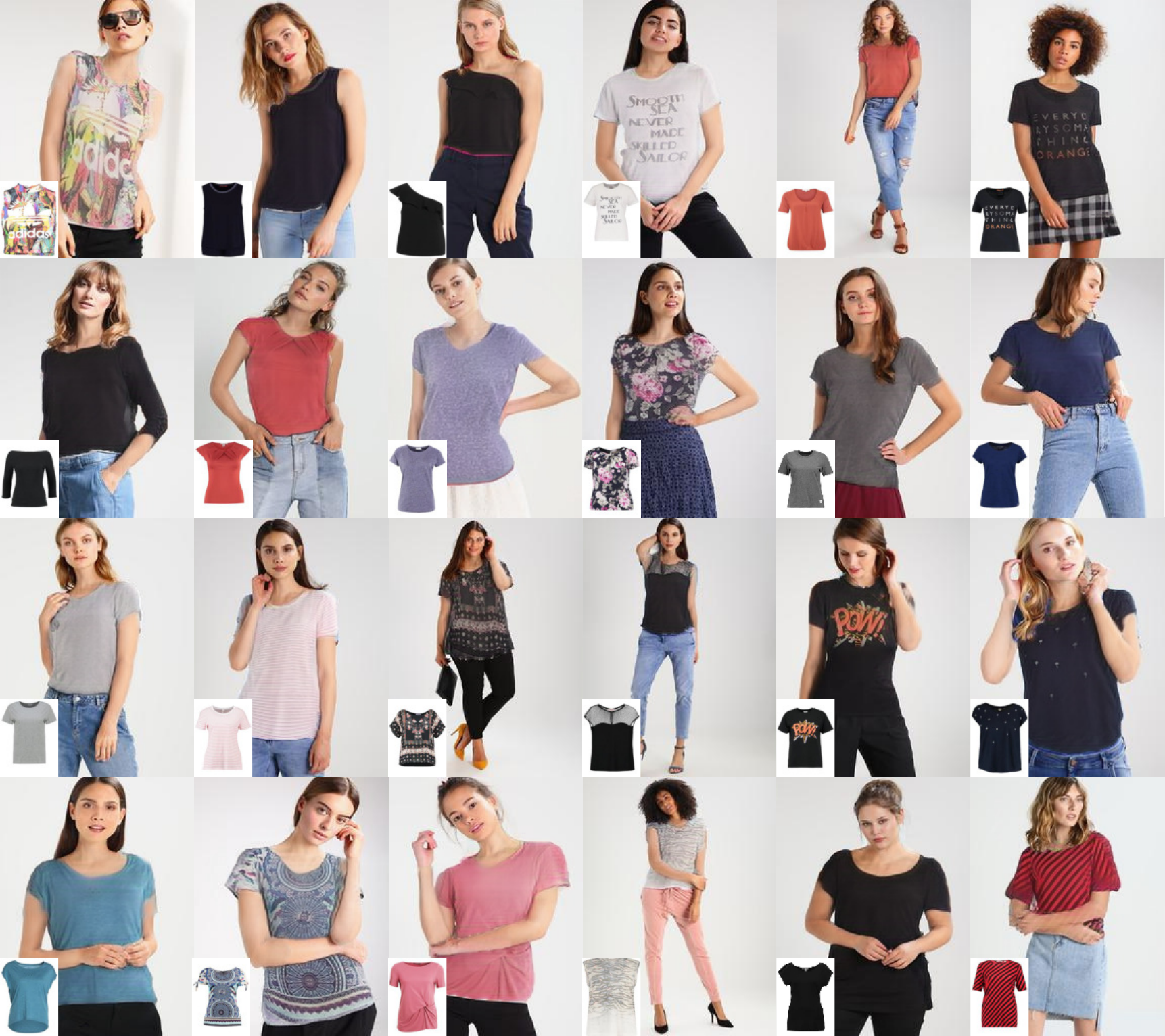}
\end{center}
   \caption{Example C-VTON results from the MPV test set containing people in what we consider difficult poses. The generated images are arranged into $4$ distinct poses to show the consistency of the synthesis procedure in a wide variety of conditions. The figure in the bottom left of each example, shows the target garment. The image is best viewed in color and electronically.}
\label{fig:appendix:mpv:pose}
\end{figure*}

To further demonstrate the capabilities of C-VTON, we present additional try-on results on the VITON and MPV datasets in Figures~\ref{fig:appendix:viton:basic} -- \ref{fig:appendix:mpv:pose}. All presented images are generated based on the test set (consisting of people and target clothing images) used throughout other parts of paper.


%The visualizations are split in the following way: 
Figures~\ref{fig:appendix:viton:basic} and \ref{fig:appendix:mpv:basic} show virtual try-on results in a basic setting, i.e. when transferring a garment without a particular texture to what we subjectively deem a simple pose\footnote{From the perspective of virtual try-on models, simple poses include arm configurations that do not substantially cover the initial clothing, so as to interfere with the segmentation/parsing procedure needed by most try-on models.}. Here, different sleeve lengths (from shortest to longest) are shown down the rows. Figures~\ref{fig:appendix:viton:texture} and \ref{fig:appendix:mpv:texture} feature the same simple pose scenario, but with textured target garments. %From top to bottom, 
The figures show transfer results with dotted patterns, followed by striped garments, smaller and larger repetitive patterns, graphics and text-based logotypes. Last but not least, Figures~\ref{fig:appendix:viton:pose} and \ref{fig:appendix:mpv:pose} show sample results in more challenging poses, ranging from poses, where arms slightly occlude the body, or are in an upward position, to poses, where the arms and hands cover a large portion of the body and corresponding clothing. %in front of the body.

Note how C-VTON is able to generate realistic results in all settings described above. Specifically, observe the quality of the warped graphics, textures and texts on the transferred clothing across all presented results and especially in  Figures~\ref{fig:appendix:viton:pose} and \ref{fig:appendix:mpv:pose}, where synthesis results with difficult poses are displayed. %We observe that in cases, there the clothing segmentation utilized to produce the masked image $I_m$, is correct, the output image $I_s$, containing a person with arms in a challenging position, is typically realistic and visually convincing.
\begin{figure*}[t]
\begin{center}
\includegraphics[width=0.95\linewidth]{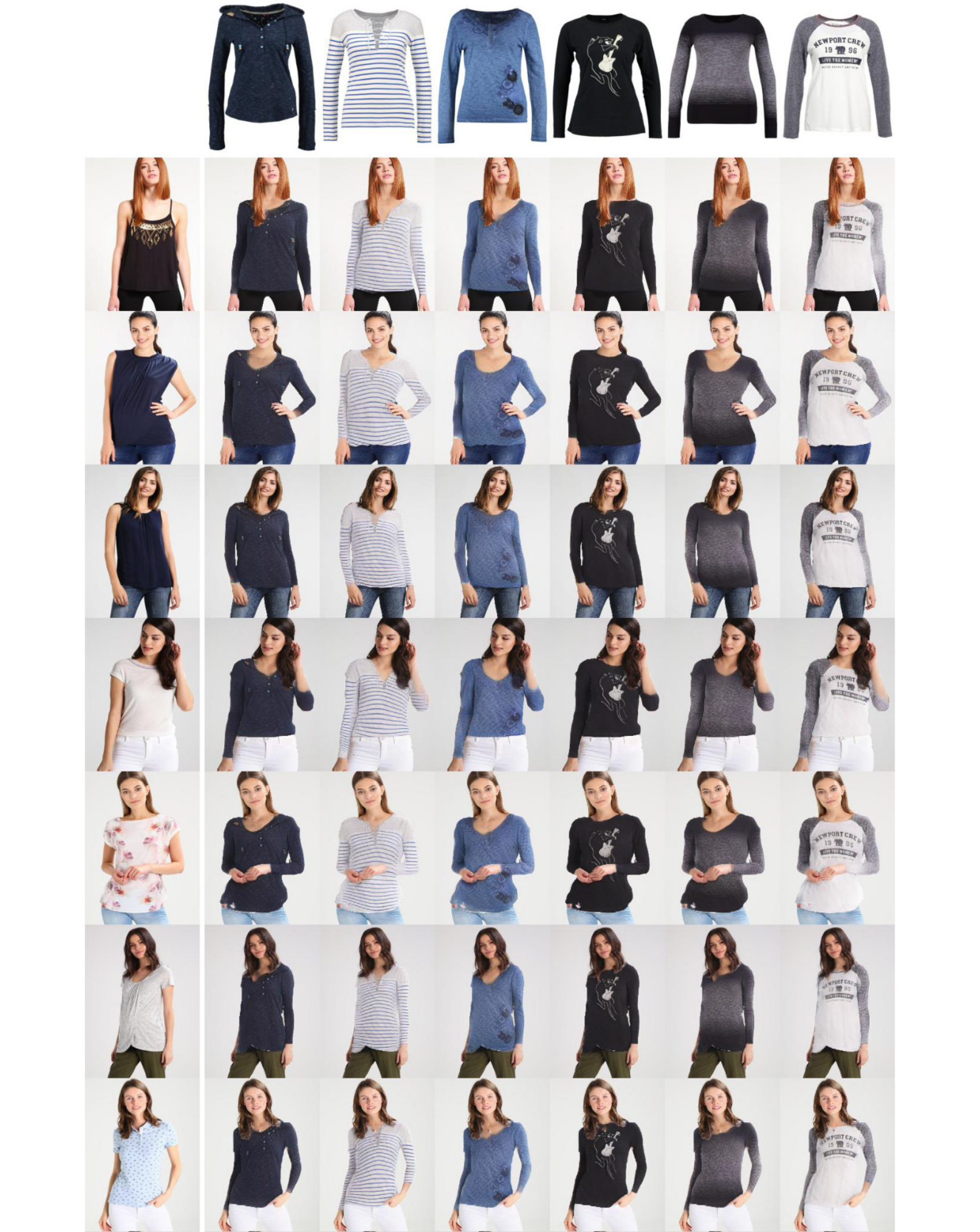}
\end{center}
   \caption{Examples of transfer from short to long-sleeved garments on the VITON dataset. The results show that realistic try-on results are achieved even with images with varying pose and/or different target clothing types. Best viewed electronically and zoomed-in.}
\label{fig:appendix:viton:short2long}
\end{figure*}
\begin{figure*}[t]
\begin{center}
\includegraphics[width=0.95\linewidth]{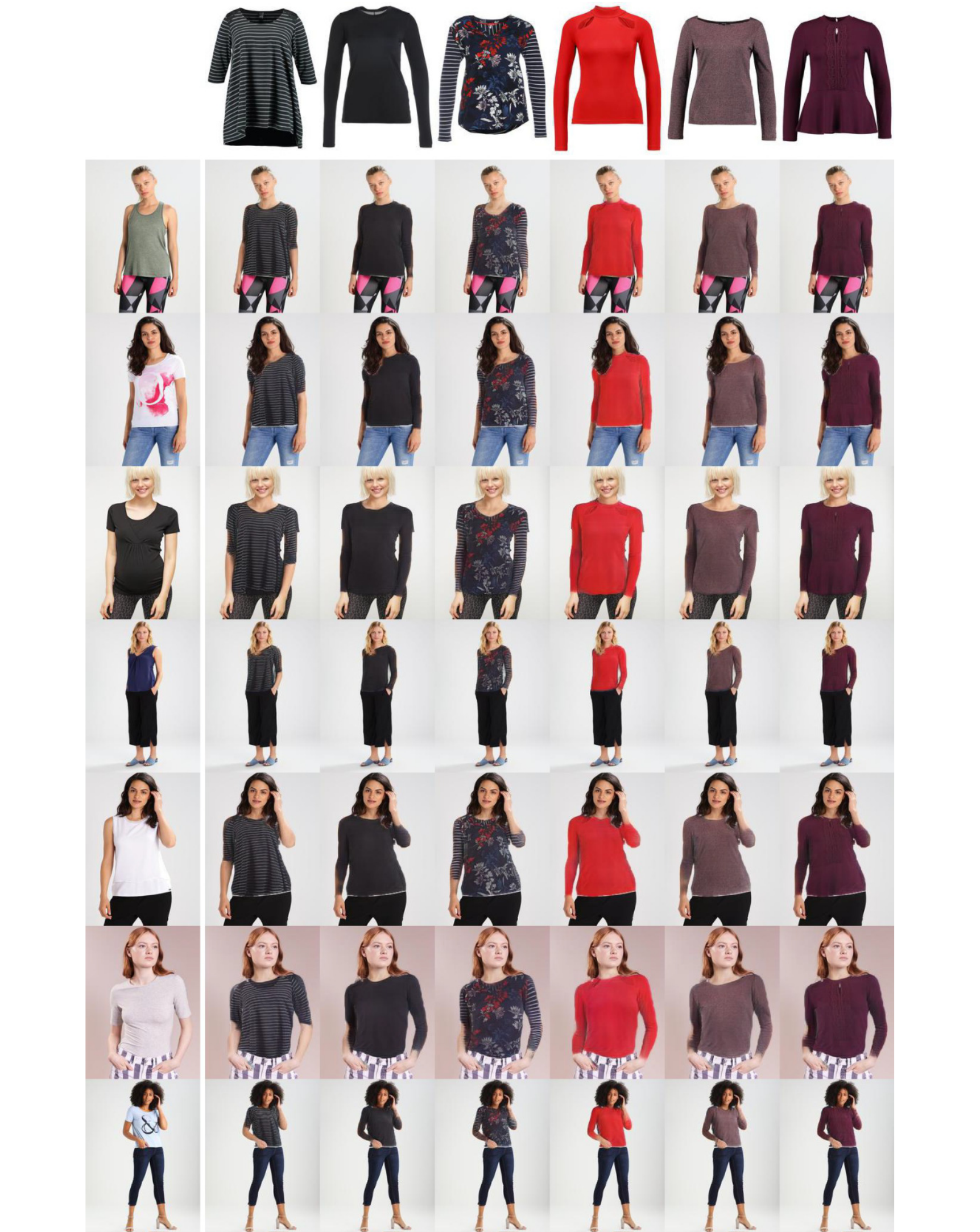}
\end{center}
   \caption{Examples of transfer from short to long-sleeved garments on the MPV dataset. The results show that realistic try-on results are achieved even with images with varying pose and/or different target clothing types. Best viewed electronically and zoomed-in.}
\label{fig:appendix:mpv:short2long}
\end{figure*}
\begin{figure*}[t]
\begin{center}
\includegraphics[width=0.95\linewidth]{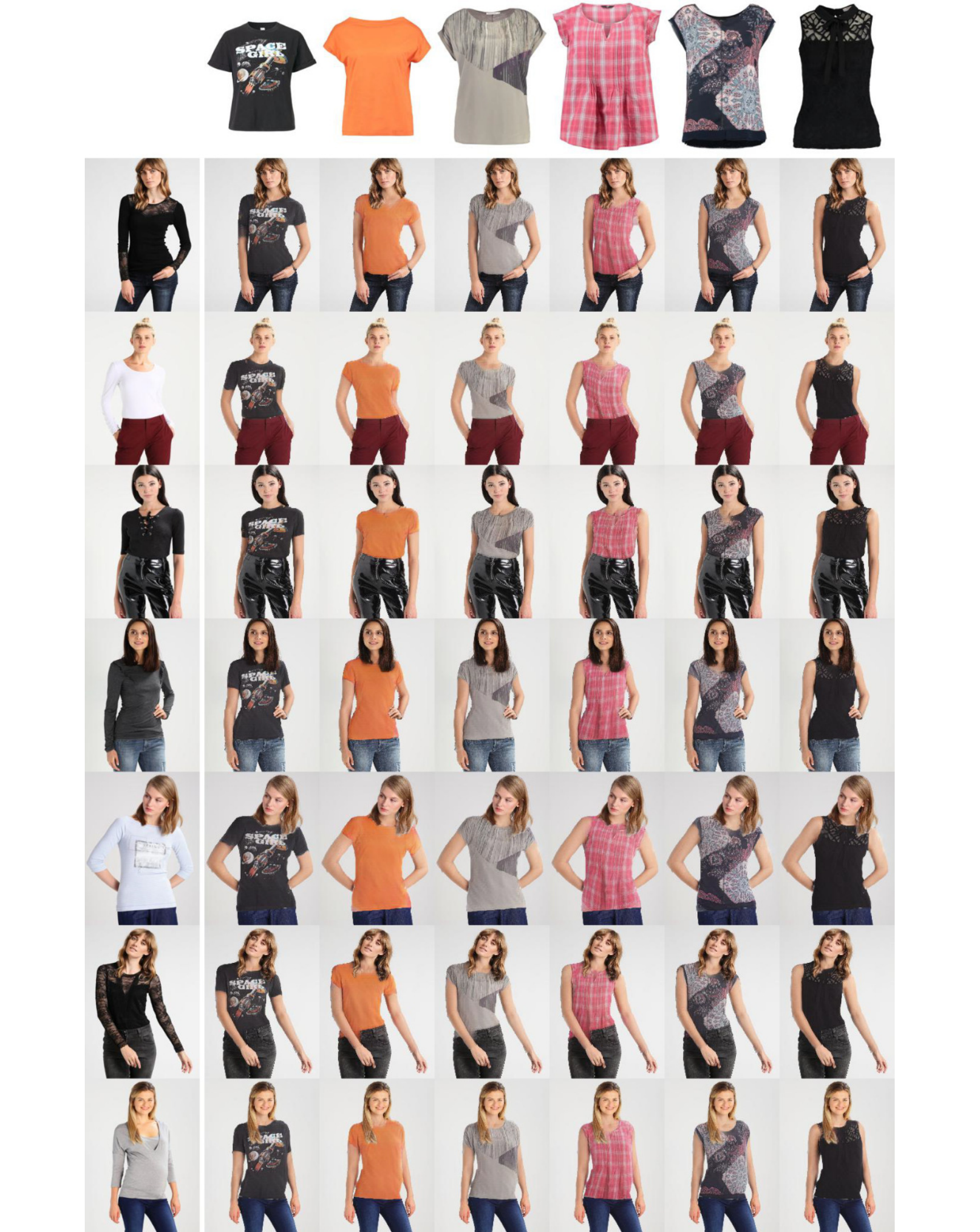}
\end{center}
   \caption{Examples of transfer from long to short-sleeved garments on the VITON dataset. The results show that realistic try-on results are achieved even with images with varying pose and/or different target clothing types. Best viewed electronically and zoomed-in.}
\label{fig:appendix:viton:long2short}
\end{figure*}
\begin{figure*}[t]
\begin{center}
\includegraphics[width=0.95\linewidth]{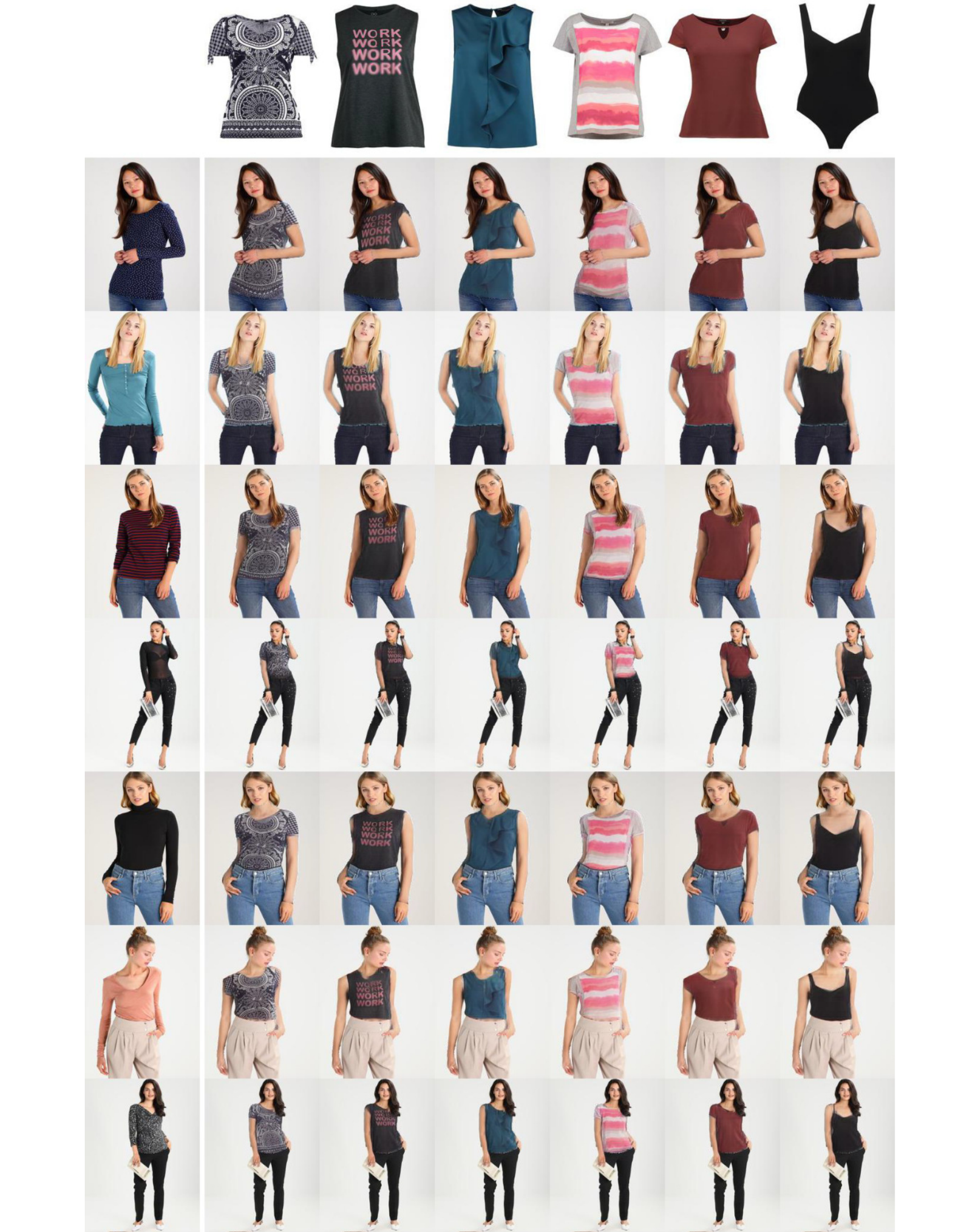}
\end{center}
   \caption{Examples of transfer from long to short-sleeved garments on MPV dataset. The results show that realistic try-on results are achieved even with images with varying pose and/or different target clothing types. Best viewed electronically and zoomed-in.}
\label{fig:appendix:mpv:long2short}
\end{figure*}
\begin{figure*}[t]
\begin{center}
\includegraphics[width=0.95\linewidth]{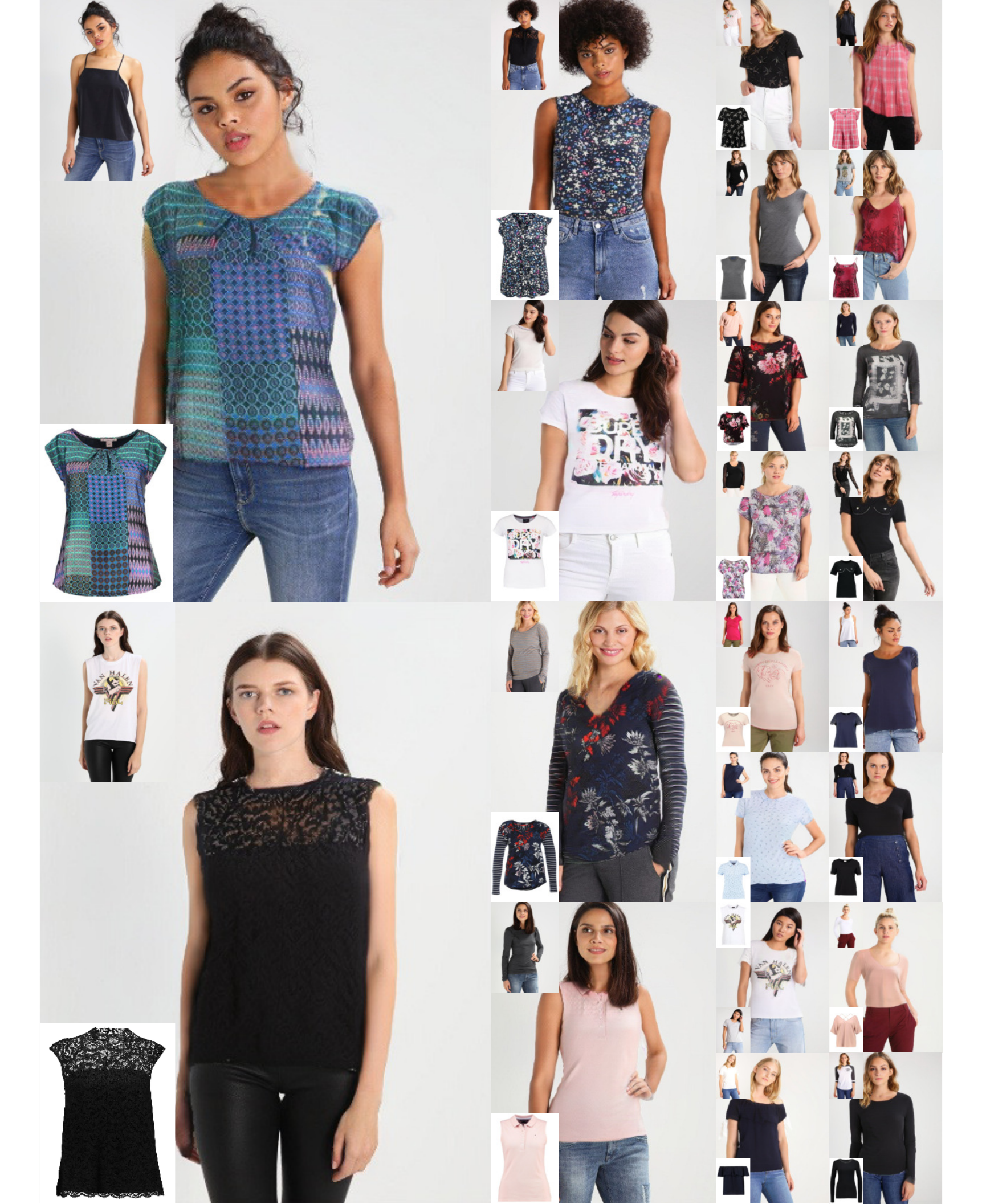}
\end{center}
   \caption{Sample results generated by C-VTON on the VITON-HD dataset. Even when synthesizing higher resolution images, C-VTON produces sharp images with comparable transfer quality to models trained on regular resolution. For each sample result presented, the image in the upper left shows the original input image, and the image in the bottom left shows the target clothing. Results are best viewed electronically.}
\label{fig:appendix:vitonhd}
\end{figure*}
\begin{figure*}[t]
\begin{center}
\includegraphics[width=\linewidth]{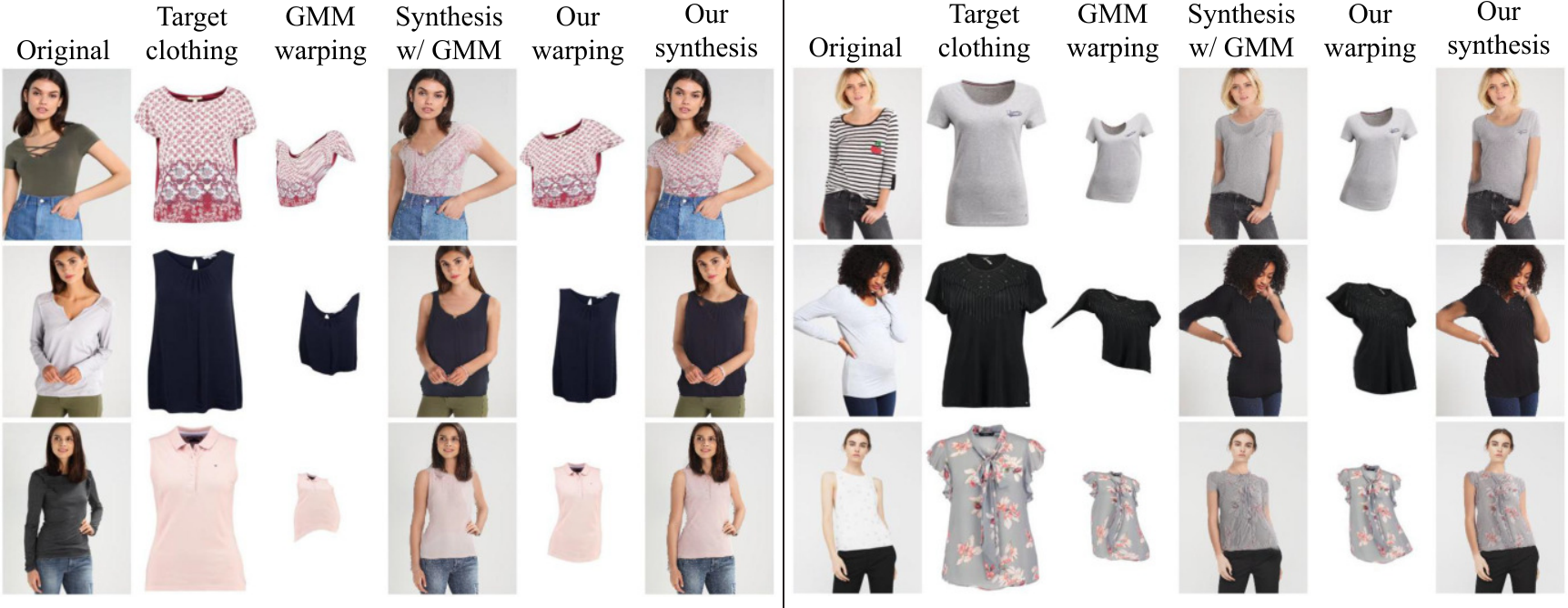}
\end{center}
   \caption{Comparison between the Geometric Matching Module from \cite{wang2018toward} and the proposed Body-Part Geometric Matcher (BPGM - marked \textit{Our warping}) on the VITON dataset. Even though C-VTON focuses on matching the body-area only, the generated results are better aligned with the person's pose.}
\label{fig:appendix:viton:warp}
\end{figure*}
\begin{figure*}[t]
\begin{center}
\includegraphics[width=\linewidth]{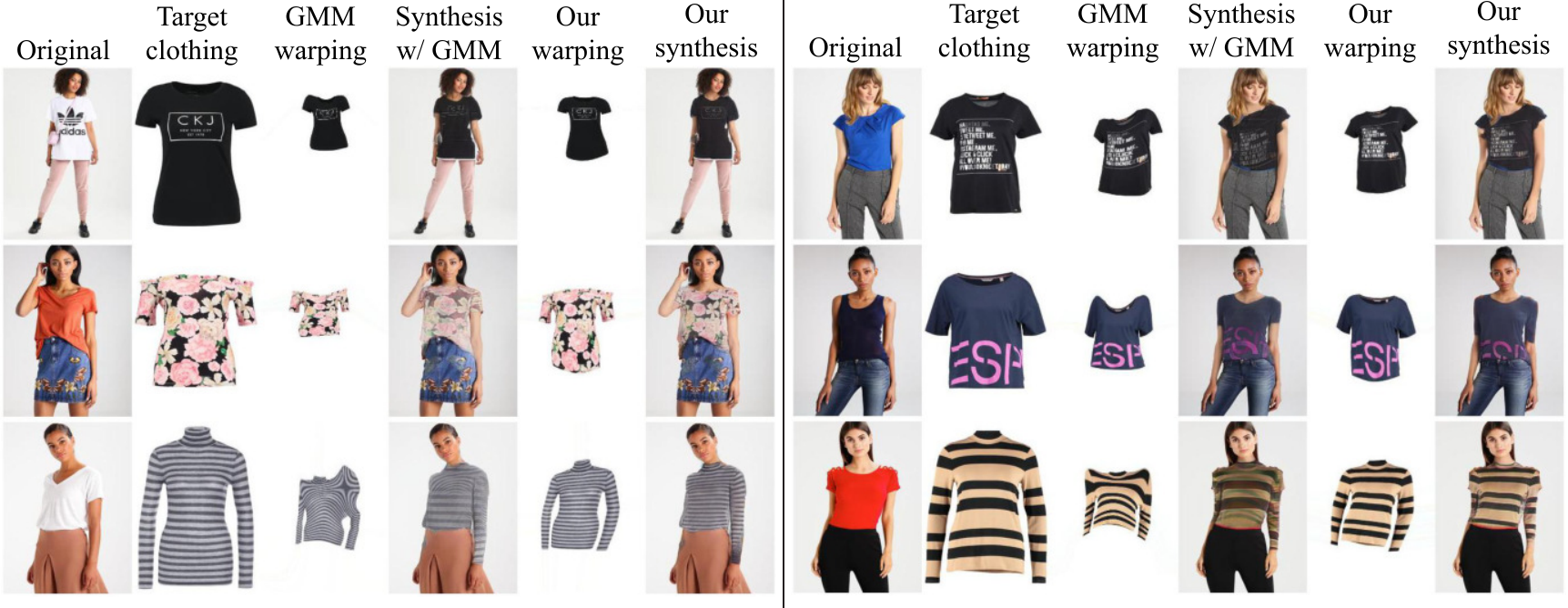}
\end{center}
   \caption{Comparison between the Geometric Matching Module from \cite{wang2018toward} and the proposed Body-Part Geometric Matcher (BPGM -  marked \textit{Our warping}) on the MPV dataset. Even though C-VTON focuses on matching the body-area only, the generated results are better aligned with the person's pose.}
\label{fig:appendix:mpv:warp}
\end{figure*}
\begin{figure*}[t]
\begin{center}
\includegraphics[width=\linewidth]{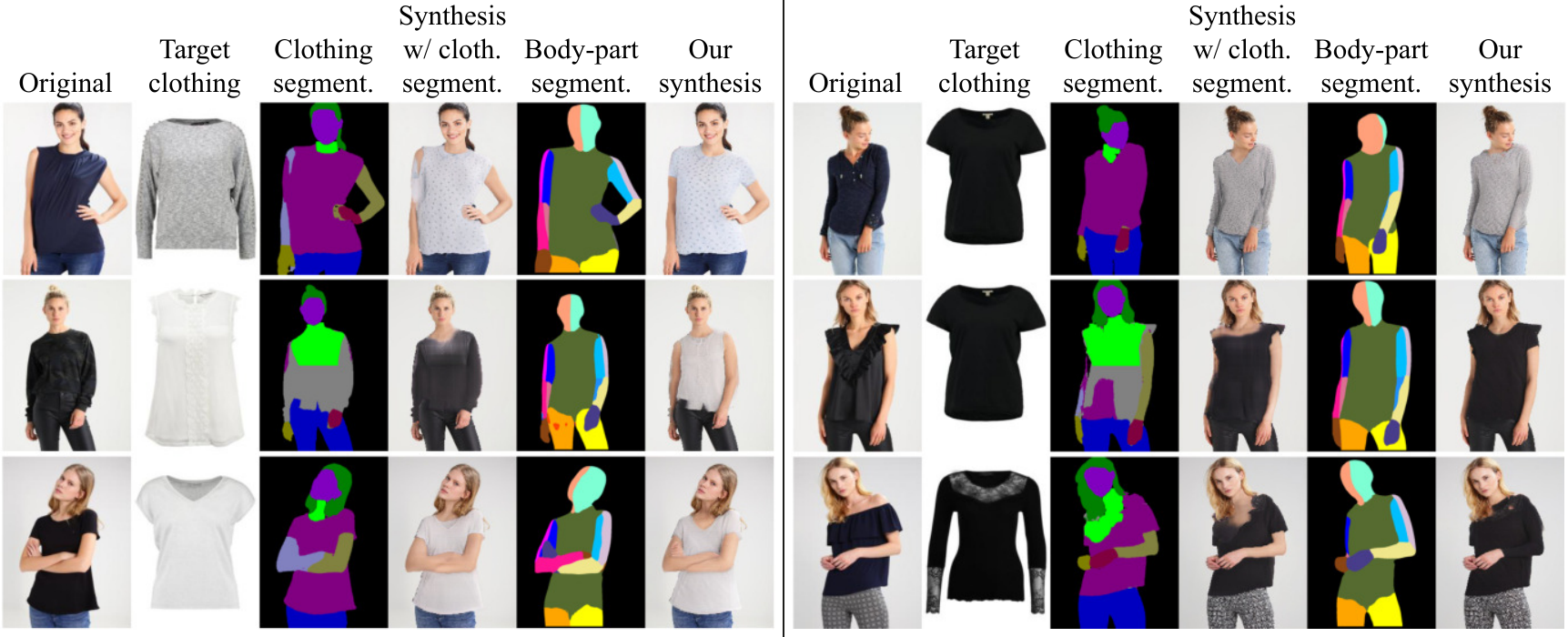}
\end{center}
   \caption{Comparison of clothing and body-part segmentations and their effect on the syntehsis quality of C-VTON on the VITON dataset. The use of clothing segmentations affects sleeve generation, garment edges and arm generation in difficult poses. The figure is best viewed in color and zoomed in.}
\label{fig:appendix:viton:seg}
\end{figure*}
\begin{figure*}[t]
\begin{center}
\includegraphics[width=\linewidth]{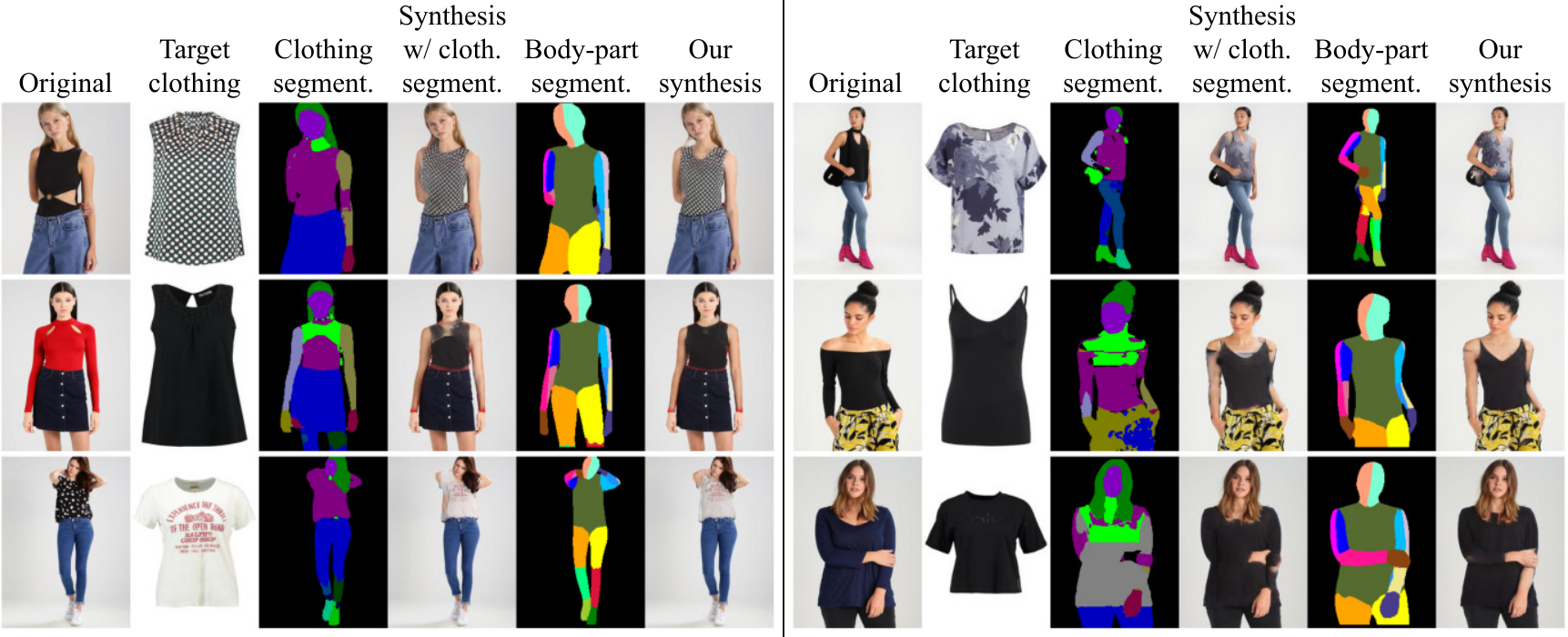}
\end{center}
   \caption{Comparisons of clothing and body-part segmentations and their effect on the syntehsis quality of C-VTON on the MPV dataset. The use of clothing segmentations affects sleeve generation, garment edges and arm generation in difficult poses. The figure is best viewed in color and zoomed in.}
\label{fig:appendix:mpv:seg}
\end{figure*}
\begin{figure*}[t]
\begin{center}
\includegraphics[width=0.95\linewidth]{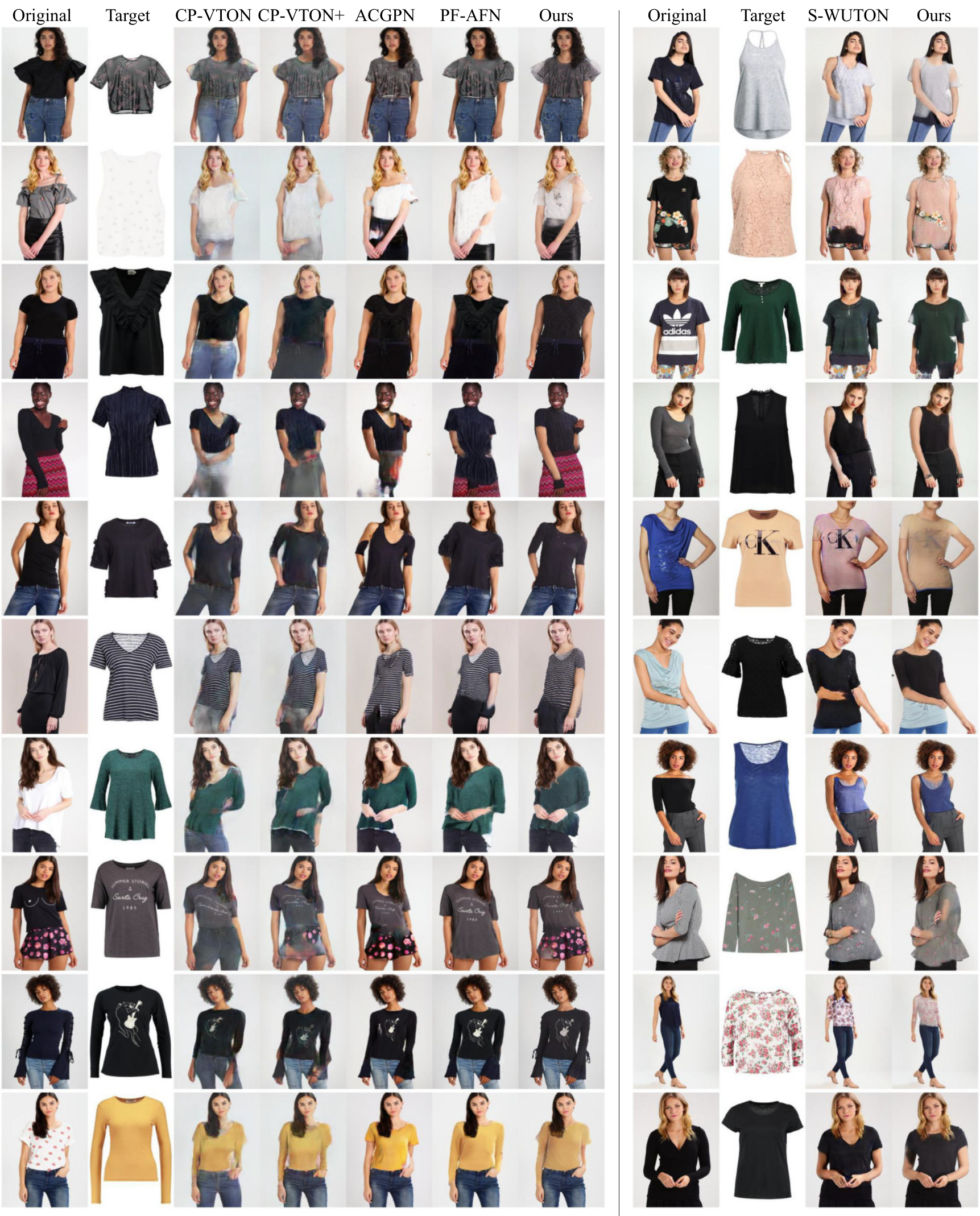}
\end{center}
   \caption{Examples of some of the limitations of C-VTON which entail blurry clothing edges and garments being only partially transferred. Note, however, that the competing methods included in the analysis have problems with realistic synthesis with the presented images as well.}
\label{fig:appendix:limit}
\end{figure*}

Next, we conduct a study involving multiple subjects and target garments. %transferring different garment types from the one that the person is originally wearing. 
The aim of the study is not only to analyze the capabilities of C-VTON when transferring the same target clothing to multiple subjects, but  to also demonstrate clothing transfers from short to long sleeves (Figures~\ref{fig:appendix:viton:short2long} and \ref{fig:appendix:mpv:short2long}) and vice versa (Figures~\ref{fig:appendix:viton:long2short} and \ref{fig:appendix:mpv:long2short}). For the study, we carefully chose a subset of test images that feature a wide range of variation with respect to arm positions, clothing that people are initially wearing and a variety of target garments with short and long sleeves. It can be seen from the presented results that C-VTON consistently generates high-quality results regardless of the target clothing and that differing sleeve lengths have little effect on the quality of the generated results. V-neck and sleeveless clothes are also largely synthesized correctly.

\section{Results on the VITON-HD Dataset} \label{appendix:vitonhd}

The results presented in the main part of the paper were generated on the VITON and MPV datasets, which contain images of size $256\times 192$ pixels. In this section, we now use a higher resolution dataset, VITON-HD, to illustrate how C-VTON performs with larger sized images. Note that VITON-HD is high-resolution version of the original VITON dataset featuring images of $512\times 368$ pixels in size and split between a training set of $14,221$ images and a testing set of $2032$ images. The dataset has only recently become popular due to being more difficult to train on. Similarly as in the main part of the paper, we again exclude all images that are present both in the training and testing data from the experimental assessment and report results over a clean test set of $416$ test images. To accommodate the higher resolution images, we slightly change the topology of the discriminators needed when training C-VTON and add $1$ additional ResNet block to the matching and patch discriminators, $D_{mth}$ and $D_{ptc}$, and $2$ additional ResNet blocks to the segmentation discriminator,  $D_{seg}$.  %a dataset desgined for training    

%Our method is also tested with VITON-HD dataset, making this approach among the first to synthesize high-resolution virtual try-on images. We used the same unpaired person and target images as with regular-sized VITON dataset and we present 
The results, presented in Figure~\ref{fig:appendix:vitonhd}, show that C-VTON can generate crisp results with minimal artefacts and generally comparable image and try-on quality as observed with models trained on the VITON and MPV datasets. In terms of quantitative scores, C-VTON achieves a FID score of $21.929$ and and LPIPS score of $0.139 \pm 0.035$ (LPIPS) over the cleaned VITON-HD test set, which is %$2.394$ and $0.031$ 
slightly worse than results for the C-VTON model trained on the VITON dataset. %Other two methods previously tested on VITON-HD dataset, PF-AFN~\cite{ge2021parser} and DCTON~\cite{ge2021disentangled}, don't have their results or models trained on this dataset publicly available, which doesn't allow us to compare our work to theirs.

\section{Impact of Design Choices} \label{appendix:effect}

We attribute a considerable part of the capabilities of C-VTON for realistic image synthesis to the use of the proposed Body-Part Geometric Matcher and reliance on body-part segmentations. To additionally showcase the importance of this contributions, we implement and train two additional C-VTON variants, i.e., 
\begin{itemize}
    \item The first uses the Geometric Matching Module (GMM) from \cite{wang2018toward} instead of the proposed Body-Part Geometric Matcher (BPGM). The change in the matching module results in differently warped clothing in the first stage of the pipeline and, therefore, has a significant impact on the final results.
    \item The second variant uses our BPGM, but relies on clothing segmentations rather than body parts (similarly in spirit to \cite{yang2020towards, yu2019vtnfp}). Because of the different correspondence, this change is again significantly affecting virtual try-on performance.
\end{itemize}%using the Geometric Matching Module (GMM) from \cite{wang2018toward} or clothing segmentations. 

Except for the changes mentioned above, the rest of the model, including the network architecture and hyperparameters used, are kept the same. Example results for the GMM versus BPGM comparison are shown in Figures~\ref{fig:appendix:viton:warp} and \ref{fig:appendix:mpv:warp}, and for the clothing versus body-part segmentations in Figures \ref{fig:appendix:viton:seg} and \ref{fig:appendix:mpv:seg}.

\textbf{GMM vs. BPGM.} When comparing warping results, it is easy to see that the GMM tries to match the garment shape precisely, leading to artefacts in the neck and arm areas. Additionally, the GMM exhibits problems with preserving on-garment textures and is susceptible to various issues, such as hair occluding the clothing the module is trying to match. Problems with on-garment texture warping can, for example, be  seen in the last row of Figure~\ref{fig:appendix:mpv:warp}, where the stripes on the garment do not match the pose of the subject well after warping. Such issues are not present in the proposed BPGM, where garments are fitted to the general body-area and not to the particular shape of the initial garment/clothing. Quantitative performance scores also point to a deterioration of results when using the GMM instead of the proposed BPGM. On VITON, the GMM-based version of C-VTON results in FID and LPIPS scores of $24.613$ and $0.136$, respectively (down by $5.078$ and $0.028$), while on MPV, FID and LPIPS scores of $6.014$ and $0.078$ are achieved (down by $1.097$ and $0.005$), respectively.

\textbf{Body-part vs. Clothing Segmentation.} A comparison between the use of body-part and clothing segmentations and their effect on the C-VTON model is shown %Contributions when using DensePose versus clothing segmentations are outlined 
in Figures~\ref{fig:appendix:viton:seg} and \ref{fig:appendix:mpv:seg}. % where each row presents a different kind of generated image quality change.
% While many other approaches[cite] synthesize their own clothing segmentations that match the target garment and use them as an additional information for the generator, we find that using such segmentations badly influences the end result. 
%In our experiment we did not synthesize clothing segmentations that match the target garment as in [TODO-cite], so the generator is not capable of swapping  differently shaped clothing (e.g. from long to short sleeve garments). Neveretheless, 
It is evident from the presented results that clothing segmentations are much noisier compared to body-part segmentations which has an adverse effect on the end results. When using clothing segmentations, the synthesized images are more susceptible to incorrect garment-swapping areas and are negatively affected by incorrect information on arm and hand areas that mostly impact image generation in difficult poses. Using body-part segmentations improves upon these issues by providing consistent information to the generator.  %, so that only negative effects of the clothing segmentations (as described in Appendix~\ref{appendix:limitations}) are present. 
Interestingly, quantitative performance scores show only minor improvements when comparing the two configurations. On VITON, FID and LPIPS scores  are $20.766$ and $0.116$ when using clothing segmentations compared to $19.535$ and $0.108$ with body-part segmentations, while on MPV these scores are $5.092$ and $0.071$ compared to $4.846$ and $0.73$ with body-part segmentations, respectively.

\section{Analysis of Limitations} \label{appendix:limitations}

Last but not least, we analyze the limitations of C-VTON and present a comparative evaluation with competing approaches in Figure~\ref{fig:appendix:limit}. As can be seen, the main issues (on both datasets) can be categorized as stemming from three different causes:
\begin{itemize}
    \setlength\itemsep{-0.1em}
    \item \textit{Preprocessing}: incorrect clothing segmentations generated during the preprocessing stage used for generating the masked input image $I_m$ that produce erroneous inputs for the context-aware generator (CAG);
    \item \textit{Input Data Characteristics}: loose clothing on the subjects in the original input image that make it difficult to infer the correct body shape and transfer the target clothing without visual (shape-induced) artefacts, and
    \item \textit{Target Clothing Characteristics}: the inability of C-VTON to identify the backside of the target garment -- especially with clothing with deeper necklines. 
\end{itemize}
These causes lead to unrealistic and soft garment edges on the synthesized images or incorrectly synthesized clothing-types and V-neck areas improperly rendered. Additionally, arms of dark-skinned people, which are the minority in the dataset, are often synthesized with lighter skin tones. %On the MPV dataset, an issue where on-garment textures lose contrast is also present. 
However, similar issues are also present with the considered competing techniques, which often result in % That being said, the methods we compare ourselves to posses similar issues that sometimes lead to 
even more unrealistic results, as illustrated by the examples in Figure~\ref{fig:appendix:limit}. For instance, most competing models also exhibit issues caused by the inability to distinguish the backside of the target clothing $C$ from the front and render sleeves inconsistently as well. The parser-free method, PF-AFN, does generate sharper clothing edges when our approach suffers from poor clothing segmentations, but nonetheless often estimates the edges incorrectly. Overall, we observe that even when C-VTON generates less convincing results, it still mostly outperforms or at least matches the quality of the results generated by the competing methods.  

\end{document}